\PassOptionsToPackage{table}{xcolor} 

\documentclass[twocolumn]{aidata}

\usepackage{amsmath}
\usepackage{array}
\usepackage{wrapfig}
\usepackage{adjustbox}
\usepackage{enumitem}
\usepackage{float}
\usepackage{pifont}
\usepackage{listings}
\usepackage{tabularx}
\usepackage{tikz}
\usepackage{pgfplots}
\usepackage{wasysym}
\usepackage{environ} 

\usepackage{amsmath,amsfonts,bm}

\def\eqref#1{equation~\ref{#1}}

\def\1{\bm{1}}

\DeclareMathAlphabet{\mathsfit}{\encodingdefault}{\sfdefault}{m}{sl}
\SetMathAlphabet{\mathsfit}{bold}{\encodingdefault}{\sfdefault}{bx}{n}

\newcommand{\gptastra}{GPT-6 Astra}
\newcommand{\gptsol}{GPT-5.6 Sol}
\newcommand{\claudefable}{Claude Fable 5}
\newcommand{\claudeopusfive}{Claude Opus 5}
\newcommand{\claudeopuseight}{Claude Opus 4.8}
\newcommand{\qwenmax}{Qwen3.8 Max}
\newcommand{\kimik}{Kimi K3}
\newcommand{\glm}{GLM-5.3}
\newcommand{\grok}{Grok-4.6}
\newcommand{\deepseek}{DeepSeek-V4-Pro}

\definecolor{taskblue}{RGB}{41, 128, 185}
\definecolor{taskpurple}{RGB}{142, 68, 173}
\definecolor{taskgreen}{RGB}{39, 174, 96}
\definecolor{taskorange}{RGB}{211, 84, 0}

\newtcolorbox{taskpanel}[2]{
  enhanced,
  colback=#1!5,
  colframe=#1!70!black,
  boxrule=0.5pt,
  arc=1.5mm,
  left=5pt, right=5pt, top=4pt, bottom=4pt,
  fonttitle=\bfseries\small,
  title=#2,
  colbacktitle=#1!15,
  coltitle=black!85
}

\lstdefinestyle{filetree}{
    basicstyle=\ttfamily\scriptsize\color{black!80},
    backgroundcolor=\color{white},
    frame=none,
    xleftmargin=0pt,
    breaklines=true,
    keepspaces=true,
    columns=fullflexible,
    showstringspaces=false
}

\definecolor{caseblue}{HTML}{356F91}
\definecolor{casegreen}{HTML}{478B68}
\definecolor{caseorange}{HTML}{C78122}
\definecolor{casepurple}{HTML}{7962A8}
\definecolor{casesoft}{HTML}{F5F7F8}

\lstdefinestyle{taskmd}{
  basicstyle=\ttfamily\scriptsize,
  breaklines=true,
  columns=fullflexible,
  keepspaces=true,
  showstringspaces=false,
  frame=single,
  rulecolor=\color{casepurple!55},
  backgroundcolor=\color{casesoft},
  xleftmargin=4pt,
  xrightmargin=4pt,
  aboveskip=5pt,
  belowskip=7pt
}

\lstdefinestyle{taskfile}{
  style=taskmd
}

\lstdefinestyle{taskpy}{
  style=taskmd,
  language=Python,
  keywordstyle=\color{caseblue}\bfseries,
  commentstyle=\color{casegreen!80!black},
  stringstyle=\color{caseorange!90!black}
}

\pgfplotsset{compat=1.18}
\usetikzlibrary{positioning,arrows.meta,fit,calc}

\newcommand{\sciswe}{\textsc{Sci-Agent-Exam}}
\definecolor{progshade}{RGB}{236,243,250}
\definecolor{rubricshade}{RGB}{240,247,242}
\definecolor{overallshade}{RGB}{246,246,246}
\definecolor{headergray}{RGB}{243,244,246}
\definecolor{scoreblue}{RGB}{61,96,132}

\newcolumntype{P}{>{\columncolor{progshade}\centering\arraybackslash}p{1.45cm}}
\newcolumntype{R}{>{\columncolor{rubricshade}\centering\arraybackslash}p{1.45cm}}
\newcolumntype{O}{>{\columncolor{overallshade}\centering\arraybackslash}p{1.45cm}}

\newcommand{\fullcircle}{\CIRCLE}
\newcommand{\halfcircle}{\LEFTcircle}
\newcommand{\emptycircle}{\Circle}

\RenewEnviron{abstract}{%
  \xdef\abstractlist{{\noexpand\abstractinfont
      \expandafter\unexpanded\expandafter{\BODY}}}%
}

\makeatletter
\if@twocolumn
  \renewcommand{\maketitle}{%
    \twocolumn[{\mymaketitle\vskip 6mm}]%
    \thispagestyle{firststyle}%
  }
\fi
\makeatother

\title{SCI-AGENT-EXAM: Benchmarking Agents on \\ Scientific Research Tasks}

\author[1,2,*]{Tianyun Zhong}
\author[3]{Wei Wang}
\author[3]{Jialin Li}
\author[4]{Jianhong Tu}
\author[5,*]{Liyao Li}
\author[4]{Chuqiao Kuang}
\author[6,*]{Tianshi Zheng}
\author[7,*]{Songchi Zhou}
\author[3]{Weiqi Zhai}
\author[3]{Qile Jin}
\author[3]{Jiajia Li}
\author[1,2]{Jiasheng Zheng}
\author[2]{Boxi Cao}
\author[2]{Yaojie Lu}
\author[2]{Hongyu Lin}
\author[1]{Shiwei Ye}
\author[3,\dagger]{Bing Zhao}
\author[3]{Hu Wei}
\author[3]{Lin Qu}
\author[2,\dagger]{Xianpei Han}

\affiliation[1]{University of Chinese Academy of Sciences}
\affiliation[2]{Institute of Software, Chinese Academy of Sciences}
\affiliation[3]{AI Data, Alibaba Group \quad $^{4}$Qwen Team, Alibaba Group}
\affiliation[5]{Zhejiang University \quad $^{6}$The Hong Kong University of Science and Technology}
\affiliation[7]{Tsinghua University}

\contribution[*]{Work done during an internship at Alibaba}
\contribution[\dagger]{Corresponding authors}

\renewcommand{\emailblock}{\email{zhongtianyun24@mails.ucas.ac.cn}, \email{xiongdao@alibaba-inc.com}, \email{xianpei@iscas.ac.cn}}

\begin{document}

\begin{abstract}

Large language model (LLM) agents are increasingly capable of supporting scientific workflows, but their capabilities across diverse research activities remain difficult to assess systematically. Existing benchmarks often focus on selected scientific capabilities or task types, limiting a comprehensive characterization of scientific agents’ strengths and weaknesses in realistic research workflows. We introduce SCI-AGENT-EXAM, a workflow-oriented benchmark comprising 120 expert-curated tasks across 11 scientific fields and 7 research task types defined by the \textsc{Scientific Workflow Task Taxonomy (SWTT)}. Each task requires both executable artifacts (e.g., code and scripts) and textual artifacts (e.g., reports and explanations). To evaluate the heterogeneous task outputs, we develop a hybrid evaluation protocol combining deterministic programmatic verification for executable artifacts with a calibrated, expert-validated LLM-based judge for textual artifacts. Beyond outcome assessment, we analyze progress and solution stability along execution trajectories to characterize how agents reach their final results. 
Experiments with 10 frontier LLM agents reveal substantial performance gaps: \gptastra{} achieves the highest overall score of 80.9, yet only 26.0\% of its runs exceed the 0.95 success threshold. Further analyses uncover distinct capability profiles across scientific task types and substantial variation in execution efficiency and stability. Our code and data are available at \url{https://github.com/SKYLENAGE-AI/Sci-Agent-Exam}.

\end{abstract}

\maketitle

\section{Introduction}



Large language models (LLMs) have evolved from systems that primarily generate and transform text into agents capable of multi-step reasoning, tool use, code execution, and interaction with external environments~\citep{yao2023react,schick2023toolformer,chen2025scienceagentbench}.
These capabilities enable LLM agents to move beyond answering individual scientific questions and participate in broader scientific workflows. 
Each scientific workflow task specifies a scientific objective,
required deliverables, and evaluation criteria, enabling assessment
under controlled execution conditions.

Evaluating scientific agents thus requires examining what they produce, whether they can execute and explain their analyses, and how reliably they reach the results.
This raises a central question: 

\textit{What can scientific agents deliver across diverse research activities, and how efficient and stable is their execution and presentation?}

Existing benchmarks have made substantial progress in evaluating scientific agents across diverse settings, including scientific question answering~\citep{wang2024scibench}, scientific discovery~\citep{majumder2024discoverybench} and analysis~\citep{chen2025scienceagentbench}, research coding~\citep{hua2025researchcodebench, siegel2024corebench}, paper replication~\citep{starace2025paperbench}, project-level research~\citep{zhou2026asibenchdawnartificialsuperintelligence}, research in a single field~\citep{laurent2024labbench} and executable scientific workflows~\citep{bragg2026astabenchrigorousbenchmarkingai,Terminal-Bench-Science_2026}.

\begin{figure}
    \centering
    \includegraphics[width=0.99\linewidth]{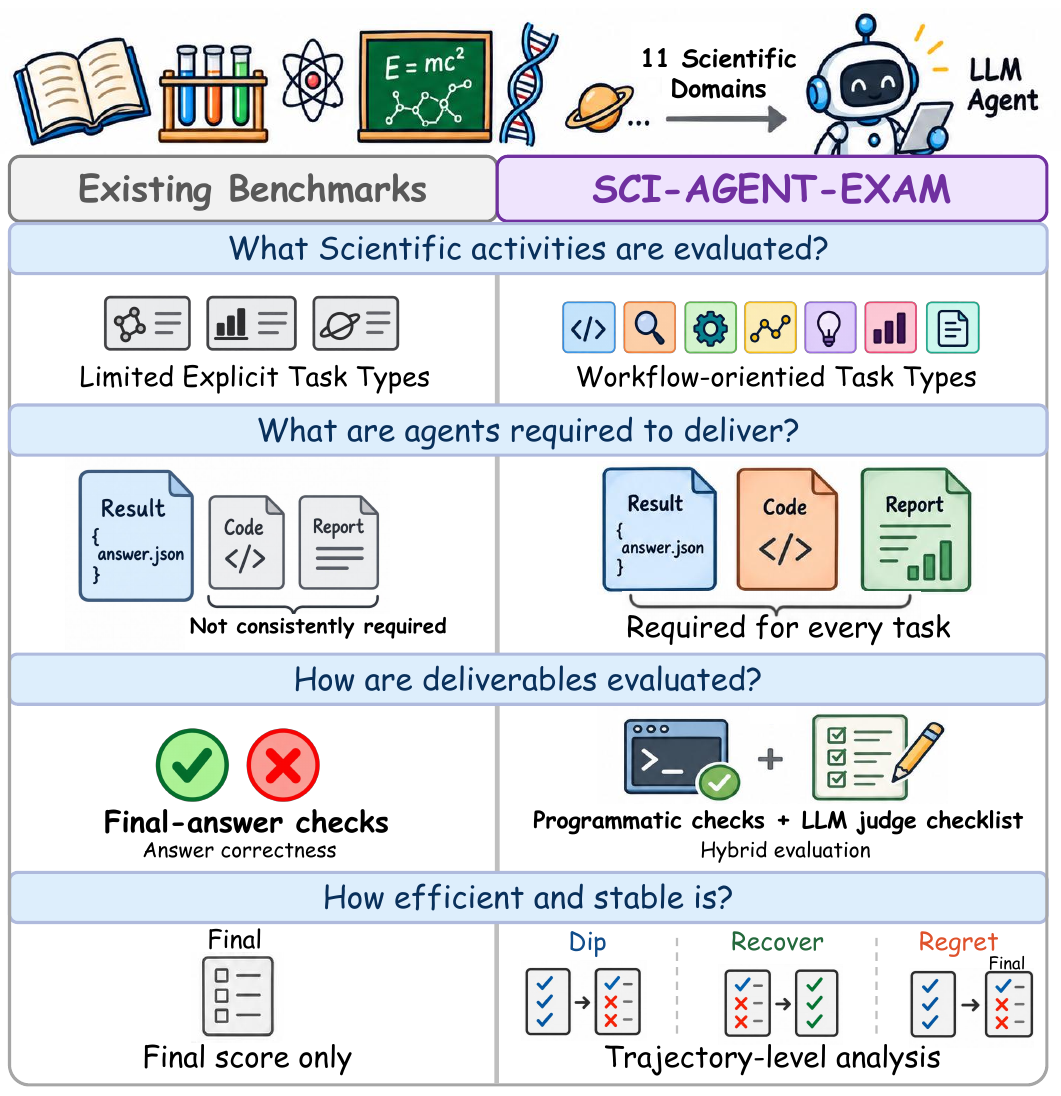}
    \caption{\sciswe{} versus existing benchmarks: workflow orientied task types, full deliverables (answer, code, report) per task, hybrid evaluation, with analysis for efficiency and stability .
    }
    \vspace{-5pt}
    \label{fig:figure1}
\end{figure}

However, as summarized in Figure~\ref{fig:figure1} and detailed in Table~\ref{tab:comparison}, two limitations remain in existing benchmarks.
First, their explicit coverage of scientific activities remains uneven, with some activities only implicitly involved, partially covered, or not covered. 
This limits a systematic characterization of agents' strengths and weaknesses across research activities. 
Second, their coverage of task deliverables remains incomplete. 
Scientific work not only requires agents to reach answers through executable results but also requires them to explain their analyses in writing. 
Yet the benchmarks compared in Table~\ref{tab:comparison} do not consistently require and jointly evaluate all three forms of deliverables (answers, executable artifacts, and textual artifacts) for every task. 
Beyond these two limitations, final-answer scores alone provide limited insight into execution efficiency and solution stability: they do not reveal how many steps agents take to reach a result or how solution quality changes during execution. 
A workflow-oriented evaluation should therefore combine systematic coverage of scientific activities with comprehensive assessment of task deliverables and analysis of execution dynamics.

\begin{table*}[t]
\centering
\fontsize{9.3}{11}\selectfont
\setlength{\tabcolsep}{1.8pt}
\renewcommand{\arraystretch}{1}

\begin{tabular}{@{}p{3.3cm}p{3.8cm}ccccccc
                @{\hspace{6pt}}c@{\hspace{4pt}}c@{\hspace{4pt}}c
                @{\hspace{6pt}}cc@{}}
\toprule
\textbf{Benchmark}
& \textbf{Task Setting}
& \multicolumn{7}{c}{\textbf{Task-Type Coverage}}
& \multicolumn{3}{c}{\textbf{Required Deliverables}}
& \multicolumn{2}{c}{\textbf{Scale}} \\
\cmidrule(lr){3-9}
\cmidrule(lr){10-12}
\cmidrule(l){13-14}
&
& \textbf{1} & \textbf{2} & \textbf{3} & \textbf{4}
& \textbf{5} & \textbf{6} & \textbf{7}
& \textbf{Answer} & \textbf{Exec.} & \textbf{Report}
& \textbf{\# Task} & \textbf{\# Domain} \\
\midrule


SciBench
& Scientific QA
& \emptycircle & \emptycircle & \emptycircle & \emptycircle
& \emptycircle & \emptycircle & \fullcircle
& \fullcircle & \emptycircle & \emptycircle
& 869 & 3 \\

DiscoveryBench
& Scientific discovery
& \emptycircle & \fullcircle & \halfcircle & \halfcircle
& \halfcircle & \emptycircle & \fullcircle
& \fullcircle & \emptycircle & \emptycircle
& 264 & 6 \\

ScienceAgentBench
& Scientific analysis
& \emptycircle & \emptycircle & \fullcircle & \fullcircle
& \halfcircle & \halfcircle & \halfcircle
& \fullcircle & \fullcircle & \emptycircle
& 102 & 4 \\

CORE-Bench
& Computational reproduction
& \emptycircle & \emptycircle & \fullcircle & \halfcircle
& \halfcircle & \halfcircle & \halfcircle
& \fullcircle & \emptycircle & \emptycircle
& 270 & 3 \\

PaperBench
& Paper replication
& \halfcircle & \emptycircle & \halfcircle & \fullcircle
& \halfcircle & \halfcircle & \halfcircle
& \emptycircle & \fullcircle & \emptycircle
& 20 & 1 \\

ResearchCodeBench
& Research coding
& \emptycircle & \emptycircle & \halfcircle & \fullcircle
& \emptycircle & \halfcircle & \emptycircle
& \emptycircle & \fullcircle & \emptycircle
& 212 & 1 \\

SWE-bench Science
& Scientific issue resolution
& \emptycircle & \emptycircle & \halfcircle & \fullcircle
& \emptycircle & \fullcircle & \emptycircle
& \emptycircle & \fullcircle & \emptycircle
& 119 & 20 \\

ASI-Bench
& Project-level research
& \halfcircle & \halfcircle & \halfcircle & \halfcircle
& \fullcircle & \halfcircle & \halfcircle
& \halfcircle & \halfcircle & \halfcircle
& 60 & 11 \\

Terminal-Bench-Science
& Scientific Tasks
& \emptycircle & \halfcircle & \halfcircle & \halfcircle
& \fullcircle & \emptycircle & \halfcircle
& \halfcircle & \halfcircle & \halfcircle
& 70 & 5 \\

\midrule
\textbf{\sciswe{}}
& \textbf{Workflow--oriented tasks}
& \fullcircle & \fullcircle & \fullcircle & \fullcircle
& \fullcircle & \fullcircle & \fullcircle
& \fullcircle & \fullcircle & \fullcircle
& \textbf{120} & \textbf{11} \\

\bottomrule
\end{tabular}
\vspace{-8pt}
\caption{
Comparison with representative scientific-agent benchmarks.
\textbf{Task-Type Coverage} maps seven task types:
\textcircled{\scriptsize 1} Literature Synthesis,
\textcircled{\scriptsize 2} Hypothesis Validation,
\textcircled{\scriptsize 3} Tool Orchestration,
\textcircled{\scriptsize 4} Code Generation,
\textcircled{\scriptsize 5} End-to-End Analysis,
\textcircled{\scriptsize 6} Error Diagnosis, and
\textcircled{\scriptsize 7} Result Interpretation.
For task-type coverage,
\fullcircle~= explicitly targeted,
\halfcircle~= involved but not distinct,
and \emptycircle~= not covered.
For \textbf{Required Deliverables},
\fullcircle~= required in every task,
\halfcircle~= required in some tasks,
and \emptycircle~= not required.
\emph{Answer}: short answer/result;
\emph{Exec.}: code/artifact;
\emph{Report}: written report.
Domain counts follow each benchmark's categorization.
}
\label{tab:comparison}
\end{table*}

\begin{figure*}[t]
    \centering
    \includegraphics[width=0.9\textwidth]{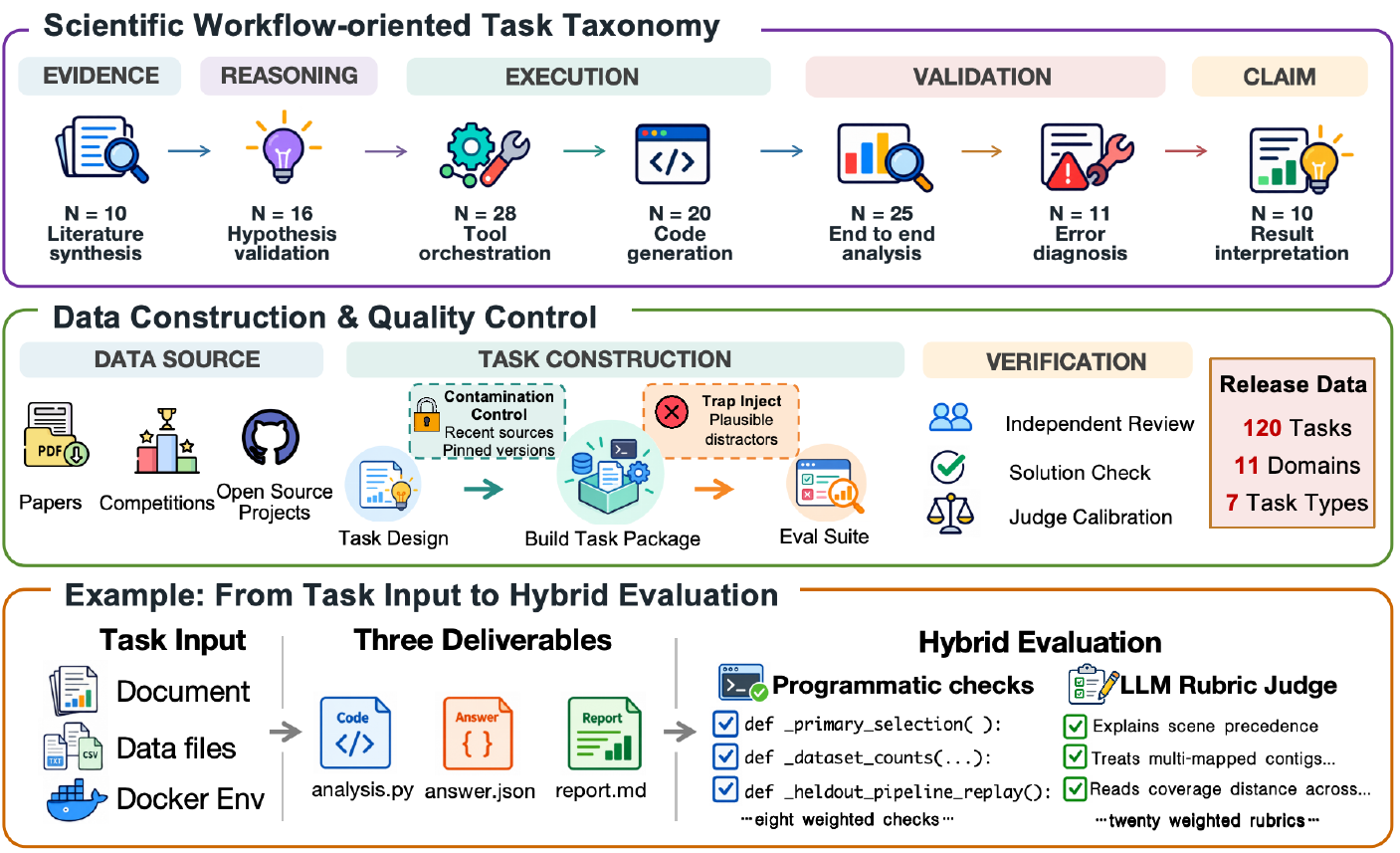}
    \vspace{-11pt}
    \caption{
    Overview of \sciswe{}. We organize tasks with a workflow-oriented taxonomy, construct executable task packages through a quality-controlled data pipeline, and evaluate task outputs using a hybrid protocol combining deterministic programmatic checks with LLM rubric judgments. The bottom panel illustrates an example task with its inputs, required deliverables, and evaluation criteria.
    }
    \vspace{-8pt}
    \label{fig:workflow}
\end{figure*}

To address these evaluation needs, we introduce \sciswe, a workflow-oriented benchmark for evaluating LLM agents in scientific research activities. First, \sciswe{} proposes the \textsc{Scientific Workflow Task Taxonomy (SWTT)}, which organizes scientific work into seven task types: Literature Synthesis, Hypothesis Validation, Tool Orchestration, Code Generation, Error Diagnosis, End-to-End Analysis, and Result Interpretation. The benchmark comprises 120 tasks spanning 11 scientific domains, with each task assigned to one task type and constructed and verified by researchers with relevant domain expertise. This taxonomy-guided design enables systematic comparison of agents' capabilities across different scientific activities.
Second, every task requires three forms of deliverables: a final structured answer, executable artifacts (e.g., code and scripts), and textual artifacts (e.g., reports and explanations). We assess these outputs using a hybrid protocol that combines deterministic programmatic verification with rubric-based LLM judgments. We additionally analyze intermediate workspace snapshots to characterize agents' execution efficiency and solution stability, including patterns of regression and recovery.

We evaluate 10 frontier LLM agents on \sciswe{} and observe substantial differences in both overall performance and task-specific capability profiles. \gptastra{} achieves the highest overall score of 80.9, outperforming the second-ranked \claudefable{} by 13.6 points. Nevertheless, only 26.0\% of its runs achieve a score of at least 0.95, indicating substantial room for improvement even for the strongest agent. Performance also varies considerably across scientific task types, revealing distinct strengths and weaknesses. Trajectory analysis further shows that stronger agents generally reach high-quality solutions earlier and undergo fewer regressions during execution.

Our main contributions are summarized as follows:
\begin{itemize}
    \item We introduce \sciswe{}, a workflow-oriented benchmark comprising 120 tasks across 11 scientific fields and 7 research task types defined by SWTT, with each task constructed and verified by PhD-level domain experts.
    \item We develop a hybrid evaluation protocol to assess both executable and textual artifacts required by each task, combining deterministic programmatic verification with a calibrated, expert-validated LLM-based judge.
    \item We evaluate 10 LLM agents, revealing uneven capabilities across scientific task types. Trajectory analysis further shows that higher-scoring agents tend to use fewer execution steps and exhibit fewer regression and recovery steps.
\end{itemize}

\section{\sciswe{}}

In this section, we introduce \sciswe{}, an executable benchmark for scientific workflow tasks. Figure~\ref{fig:workflow} summarizes the task taxonomy,
construction pipeline, and hybrid evaluation protocol of
\sciswe{}, an executable benchmark for scientific workflow tasks.


\subsection{Task formulation and taxonomy}

Each task specifies a scientific objective, execution environment,
visible resources, required deliverables, and evaluation criteria,
enabling assessment under controlled execution conditions.

We represent task $i$ as
\begin{equation}
    \mathcal{T}_i = (I_i, E_i, V_i, D_i, C_i, R_i),
\end{equation}
where $I_i$ denotes the scientific instruction, $E_i$ the fixed execution
environment, $V_i$ the files and source snapshots visible to the agent, and
$D_i$ the required deliverable specification. The evaluator additionally holds
deterministic checks $C_i$ for executable artifacts and a human-written rubric
$R_i$ for textual artifacts. Given $(I_i,E_i,V_i,D_i)$, the agent interacts
with the environment and available tools to produce an artifact set
$\hat{A}_i$. Evaluator-held checks and references remain hidden from the agent,
allowing the resulting outputs to be assessed under controlled conditions.


\textbf{Scientific Workflow Task Taxonomy.} We organize scientific research tasks into seven categories,
collectively termed the \textit{Scientific Workflow Task Taxonomy}
(SWTT).

As shown in Table~\ref{tab:domain_type_distribution}, SWTT covers seven task types:
(1) \textit{Literature Synthesis}, which requires reconciling evidence from papers or source snapshots into a grounded recommendation;
(2) \textit{Hypothesis Validation}, which requires testing a scientific hypothesis against alternative explanations and producing a calibrated verdict;
(3) \textit{Tool Orchestration}, which requires composing multiple tools or data sources while correctly propagating dependencies, schemas, units, and intermediate results;
(4) \textit{Code Generation}, which requires producing executable code that satisfies a specified interface and output contract;
(5) \textit{End-to-End Analysis}, which requires carrying out a complete analysis from data and method selection to validation and qualified conclusions;
(6) \textit{Error Diagnosis}, which requires identifying and minimally repairing a runnable project that produces scientifically incorrect results;
(7) \textit{Result Interpretation}, which requires translating experimental or statistical results into claims whose scope and limitations are supported by the available evidence.

 We cross the task types with 11 scientific domains. The matrix includes physical sciences, physics simulation, mathematics, materials science, computer science, engineering, chemistry, data science, biology, scientific computation and bioinformatics.

\subsection{Benchmark Construction}

Our pipeline comprises data sourcing, task construction,
verification, and release (Figure~\ref{fig:workflow}).
A total of 156 domain experts participated, with task production
and evaluation conducted by independent groups.

\textbf{Data sourcing.}
We draw tasks from recent research papers, scientific competitions
and datasets, and actively maintained open-source projects,
selecting problems that can be executed and independently verified.
Source-specific contamination controls and selection criteria
are detailed in Appendix~\ref{app:task-package}.

\textbf{Task construction.}
Each selected source is transformed by a PhD-level domain expert through three
steps: \textit{task design}, \textit{task package construction}, and
\textit{evaluation suite construction}. First, the expert defines the
scientific objective, evidence requirements, constraints, and expected
deliverables. Second, the expert builds a self-contained executable package
containing the task instruction, input data, supporting resources, software
environment, and required tools, together with a reference solution that
establishes task feasibility. Where appropriate, \textit{trap injection}
introduces plausible distractors or misleading intermediate signals that
require the agent to distinguish scientifically valid evidence from shortcut
solutions. Third, the expert constructs the evaluation suite, including
deterministic programmatic checks for executable outputs and human-written
rubric criteria for scientific presentation. The rubric is written after the
reference solution rather than generated by an LLM, and the reference solution
must achieve at least 90\% on the corresponding rubric as a calibration check.
Writers targeted approximately six hours per task across design,
implementation, and rubric construction. Detailed trap cases are provided in Appendix~\ref{app:task-package}.


\textbf{Verification.}
Each completed task then passes through the three verification steps shown in
Figure~\ref{fig:workflow}: \textit{independent review}, \textit{solution check},
and \textit{judge calibration}. Independent experts review the scientific
validity and task specification; the reference workflow is rerun to verify
executability and the correctness of expected outputs; and the evaluation
criteria are checked for alignment with the task objective and reference
solution. Data production and evaluation are performed by different experts,
and a task is released only after passing two independent expert stages.
Reviewers and super-reviewers targeted approximately two hours and one hour per
task, respectively, with reviewer--super-reviewer agreement subject to an 85\%
quality threshold.

\begin{table}[t]
\centering
\small
\setlength{\tabcolsep}{2.5pt}
\renewcommand{\arraystretch}{1.08}
\begin{tabular}{@{}lrrrrrrr@{\hspace{5pt}}r@{}}
\toprule
\textbf{Domain}
& \textcircled{\scriptsize 1}
& \textcircled{\scriptsize 2}
& \textcircled{\scriptsize 3}
& \textcircled{\scriptsize 4}
& \textcircled{\scriptsize 5}
& \textcircled{\scriptsize 6}
& \textcircled{\scriptsize 7}
& \textbf{Total} \\
\midrule
Bioinformatics         & 0 & 1 & 2 & 1 & 5 & 2 & 0 & 11 \\
Biology                & 1 & 2 & 4 & 2 & 4 & 1 & 1 & 15 \\
Chemistry              & 3 & 1 & 3 & 2 & 1 & 1 & 0 & 11 \\
Computer science       & 0 & 3 & 1 & 3 & 2 & 1 & 0 & 10 \\
Data science           & 1 & 2 & 2 & 1 & 2 & 1 & 1 & 10 \\
Engineering            & 2 & 0 & 2 & 1 & 1 & 0 & 2 &  8 \\
Materials science      & 0 & 2 & 3 & 1 & 1 & 1 & 2 & 10 \\
Mathematics            & 1 & 0 & 1 & 2 & 2 & 2 & 1 &  9 \\
Physical science       & 0 & 1 & 3 & 3 & 2 & 1 & 2 & 12 \\
Physics simulation     & 2 & 2 & 3 & 2 & 2 & 0 & 0 & 11 \\
Scientific computation & 0 & 2 & 4 & 2 & 3 & 1 & 1 & 13 \\
\midrule
\textbf{Total}
& \textbf{10} & \textbf{16} & \textbf{28} & \textbf{20}
& \textbf{25} & \textbf{11} & \textbf{10} & \textbf{120} \\
\bottomrule
\end{tabular}
\caption{Distribution of 120 tasks across 11 scientific domains
and seven SWTT task types. Column numbers follow the definitions
in the text and Table~\ref{tab:comparison}.}
\label{tab:domain_type_distribution}
\end{table}

\subsection{Hybrid evaluation protocol}
\label{sec:evaluation}

\begin{table*}[t]
\centering
\small
\begin{tabular}{clcccc}
\toprule
\textbf{Rank} & \textbf{Model}
& \textbf{Programmatic}
& \textbf{LLM Judge}
& \textbf{Final}
& \textbf{Pass@0.95} \\
\midrule
1
& \gptastra{}
& $\mathbf{77.7}\,{\scriptsize \pm 2.5}$
& $\mathbf{85.3}\,{\scriptsize \pm 1.4}$
& $\mathbf{80.9}\,{\scriptsize \pm 1.7}$
& $\mathbf{26.0}$ \\

2
& \claudefable{}
& $\underline{62.7}\,{\scriptsize \pm 3.1}$
& $\underline{74.4}\,{\scriptsize \pm 2.0}$
& $\underline{67.3}\,{\scriptsize \pm 2.0}$
& $9.7$ \\

3
& \claudeopusfive{}
& $59.2\,{\scriptsize \pm 2.8}$
& $71.8\,{\scriptsize \pm 2.4}$
& $64.2\,{\scriptsize \pm 2.1}$
& $\underline{10.5}$ \\

4
& \qwenmax{}
& $54.7\,{\scriptsize \pm 2.6}$
& $74.2\,{\scriptsize \pm 1.6}$
& $62.7\,{\scriptsize \pm 1.7}$
& $7.4$ \\

5
& \kimik{}
& $57.6\,{\scriptsize \pm 2.6}$
& $71.2\,{\scriptsize \pm 1.6}$
& $62.6\,{\scriptsize \pm 1.7}$
& $6.1$ \\

6
& \gptsol{}
& $58.8\,{\scriptsize \pm 2.5}$
& $62.1\,{\scriptsize \pm 1.8}$
& $59.6\,{\scriptsize \pm 1.8}$
& $6.9$ \\

7
& \claudeopuseight{}
& $54.6\,{\scriptsize \pm 2.5}$
& $63.5\,{\scriptsize \pm 1.8}$
& $57.1\,{\scriptsize \pm 1.7}$
& $3.2$ \\

8
& \glm{}
& $47.7\,{\scriptsize \pm 2.7}$
& $67.6\,{\scriptsize \pm 2.4}$
& $55.9\,{\scriptsize \pm 2.1}$
& $5.3$ \\

9
& \grok{}
& $49.0\,{\scriptsize \pm 3.0}$
& $57.4\,{\scriptsize \pm 2.7}$
& $52.4\,{\scriptsize \pm 2.4}$
& $3.1$ \\

10
& \deepseek{}
& $44.8\,{\scriptsize \pm 2.7}$
& $61.1\,{\scriptsize \pm 1.8}$
& $50.9\,{\scriptsize \pm 1.6}$
& $1.9$ \\
\bottomrule
\end{tabular}
\vspace{-8pt}
\caption{
Model performance on \sciswe{} (120 tasks; Claude Code harness).
Programmatic and LLM Judge scores assess executable and textual artifacts,
respectively. Scores are mean $\pm$ standard error across tasks;
Pass@0.95 is the fraction of runs with a final score of at least 0.95.
Bold and underlined values mark the best and second-best results.
}
\label{tab:main}
\end{table*}

Open-ended scientific tasks yield heterogeneous artifacts and claims that no single paradigm can judge well~\citep{liang2023holisticevaluationlanguagemodels,liu-etal-2023-g}. We therefore decompose each task into discrete \emph{scoring points} and grade them through two complementary channels that are reported separately before being combined. Every scoring point carries an importance weight set during annotation: a programmatic check realizes it as an executable \emph{function}, while the rubric realizes it as a pass/fail \emph{criterion}.

\textbf{Controlled execution.}
Every task runs in a dedicated Docker container with a fixed scientific stack and task-specific tools, and all models act through the same command-line harness. The harness builds the image, invokes the agent, collects the produced files, runs the deterministic checker on hidden inputs, applies the judge, and writes an evaluation receipt---fixing the interface and execution conditions while leaving the analysis strategy to the agent.

\textbf{Programmatic score.}
Each scoring point is implemented as a checker function $c_{ik}\in[0,1]$ with weight $\alpha_{ik}$, giving
\begin{equation}
p_i=\frac{\sum_{k=1}^{K_i}\alpha_{ik}c_{ik}}{\sum_{k=1}^{K_i}\alpha_{ik}}.
\end{equation}
A function imports the generated code or reads the required artifact and reruns the computation on hidden data~\citep{chen2025scienceagentbench,siegel2024corebench}, testing execution status, numerical values within declared tolerances, output schemas, required files, invariance under allowed perturbations, and task-specific constraints. A missing artifact scores zero on every function that depends on it.

\textbf{Rubric-based judgment.}
Functions cannot tell whether a literature comparison is faithful, a causal claim is calibrated, or a limitation is scientifically material~\citep{liu-etal-2023-g}. Each remaining scoring point is therefore written as a rubric criterion $q_{ir}\in\{0,1\}$ with weight $\beta_{ir}$, spanning scientific correctness, execution and validation, source grounding, communication, and completeness~\citep{chen2025scienceagentbench,zheng2023judging}, giving
\begin{equation}
j_i=\frac{\sum_{r=1}^{R_i}\beta_{ir}q_{ir}}{\sum_{r=1}^{R_i}\beta_{ir}}.
\end{equation}
Preflight marks every criterion that depends on an absent file as failed rather than inferring it from the final prose; a malformed or unavailable judge is logged as a setup failure and excluded from scoring. The full criterion inventory and calibration procedure appear in Appendix~\ref{app:full-case-rubric}.

\textbf{Score aggregation.}
The two scores are combined as
\begin{equation}
s_i = w_i^p\, p_i + w_i^j\, j_i, \qquad w_i^p+w_i^j=1,
\end{equation}
with the weights fixed per task type before evaluation. The split $(w_i^p, w_i^j)$ is not uniform but varies with the task type. We always report $p_i$ and $j_i$ beside $s_i$.
The per-task-type weights are listed in Table~\ref{tab:task_type_weights} of the appendix.

\section{Experiments}

\begin{figure*}[t]
    \centering
    \includegraphics[width=\textwidth]{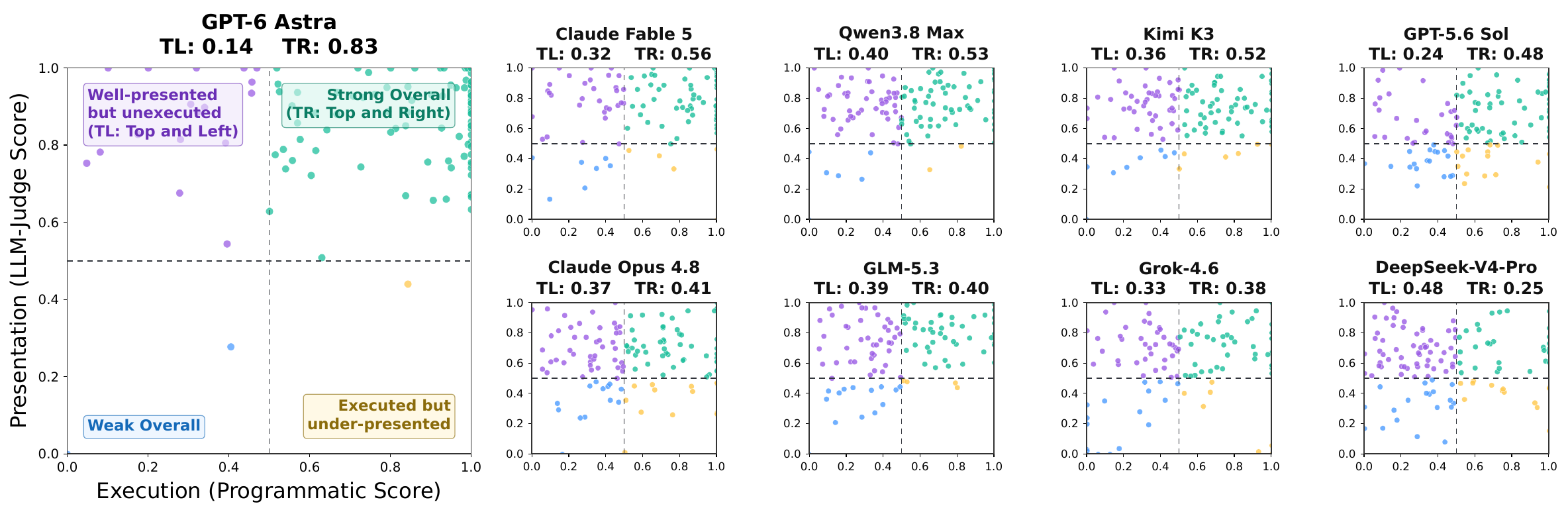}
    \vspace{-13pt}

\caption{
\textbf{
Agents generally report better than they execute.}
Task-level outcomes frequently concentrate in the top-left (TL) quadrant,
indicating strong report quality but weak executable correctness.
TL is the fraction of cases with LLM-judge $\geq 0.5$ and Programmatic
$< 0.5$; TR is the fraction for which both scores are $\geq 0.5$.
}
    \vspace{-4pt}
    \label{fig:prog_vs_judge}
\end{figure*}




\begin{figure*}[t]
  \centering


    \centering
    \includegraphics[width=0.9\textwidth]{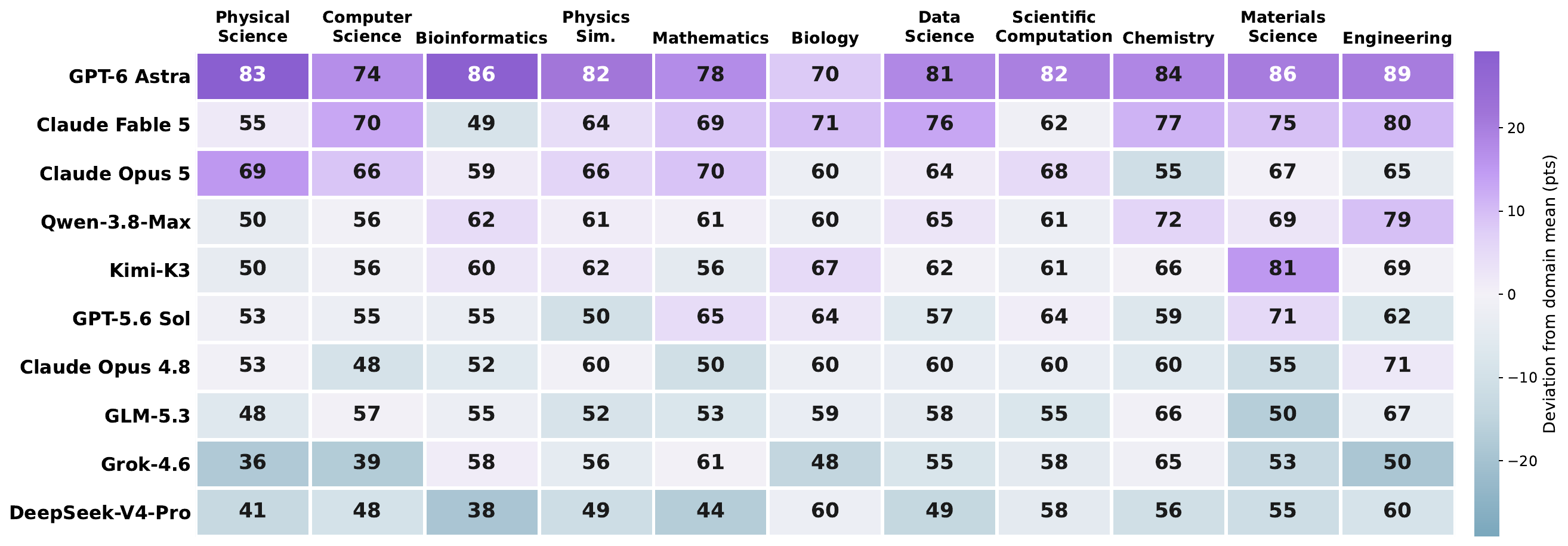}
  \caption{
\textbf{Model performance across scientific domains.} Rows are ordered by overall benchmark performance.
}
  \label{fig:heatmap_domains}
\end{figure*}

We evaluate scientific agents from three complementary perspectives.
First, we examine final performance in executable correctness and
rubric-based scientific quality, and analyze how performance varies across
scientific domains and task types. Second, we study execution trajectories
to characterize progress and solution stability, including regression and
recovery behaviors. Third, we examine execution efficiency and cost to
understand the resources agents require to achieve their results.


\subsection{Experimental setup}

\textbf{Agent configuration.}
We use Claude Code as the common agent harness and set the inference effort
to \texttt{high} for all models. Each model is run independently three times
per task under the same visible task package and execution protocol.
We examine sensitivity to the agent harness and inference-effort setting in
Appendix~\ref{app:evaluation_robustness}.



\textbf{Models.}
We evaluate ten models spanning proprietary and open-weight model families. The proprietary model families include \gptastra{}~\citep{openai2026gpt6}, \gptsol{}~\citep{openai2025gpt5}, \claudeopusfive{}~\citep{claudeopus5}, \claudefable{}~\citep{claudefable5}, \claudeopuseight{}~\citep{claudeopus48} and \grok{}~\citep{grok46}. The open-weight model families include \kimik{}~\citep{kimiteam2026kimi}, \qwenmax{}~\citep{qwen38}, \glm{}~\citep{glm5team2026glm5vibecodingagentic}, and \deepseek{}~\citep{deepseekai2026deepseekv4}.


\textbf{Evaluation procedure.}
Each task has approximately 8 programmatic checks and 20 LLM-judge
rubric criteria. For each run, we preserve the generated artifacts,
score the programmatic checks, and use \gptsol{}~\citep{openai2025gpt5} to assess the rubric criteria. These yield a programmatic score $p_i$, an LLM-judge score $j_i$, and a task-weighted final score $s_i$. We first average the three runs for each model then report macro averages across tasks.



\subsection{Main Results}

\begin{figure*}[t]
\centering
\begin{subfigure}[t]{0.36\textwidth}
    \centering
    \includegraphics[width=\linewidth]{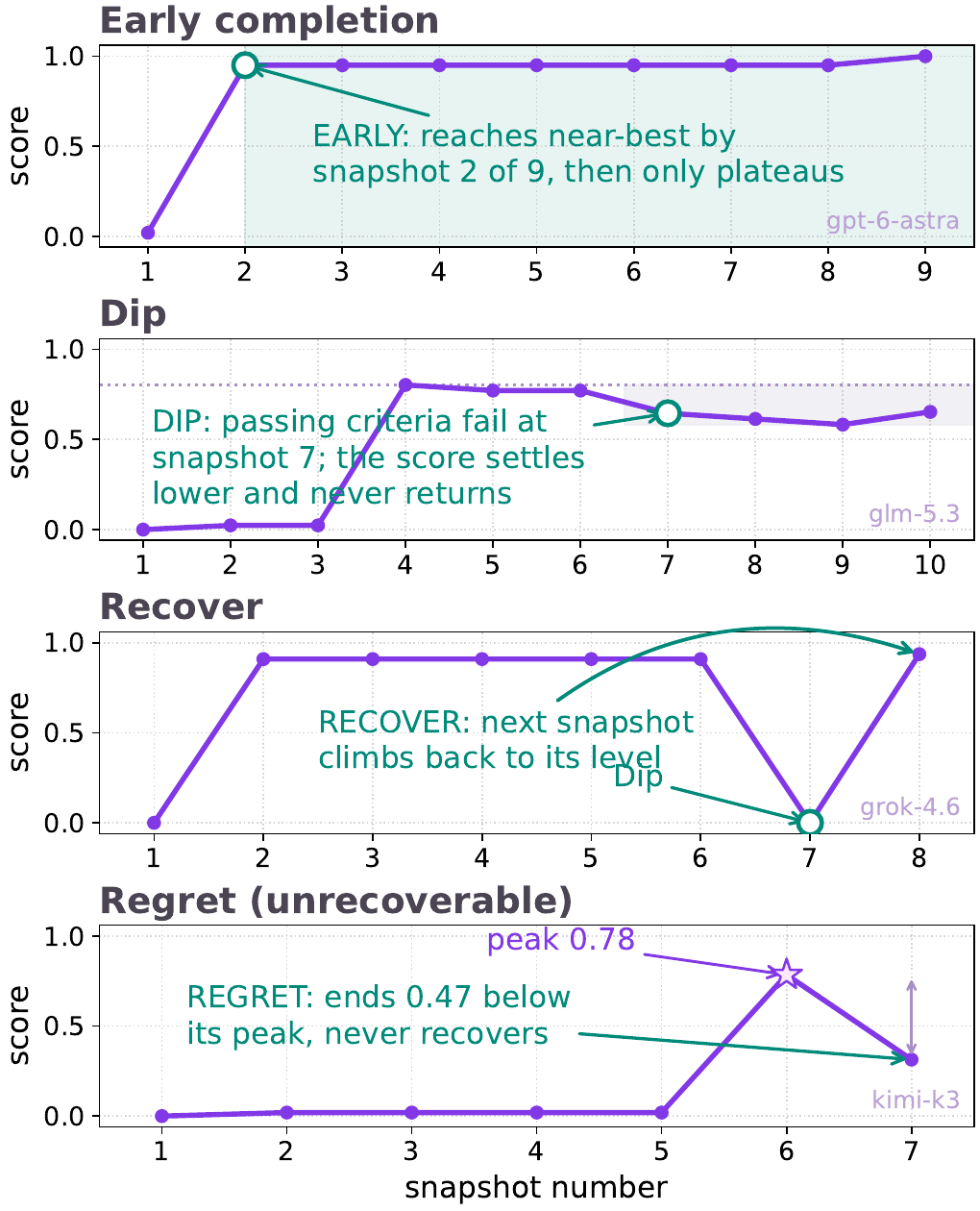}
    \caption{Four examples illustrative trajectory types.}
    \label{fig:behaviors_def}
\end{subfigure}
\hfill
\begin{subfigure}[t]{0.6\textwidth}
    \centering
    \includegraphics[width=\linewidth]{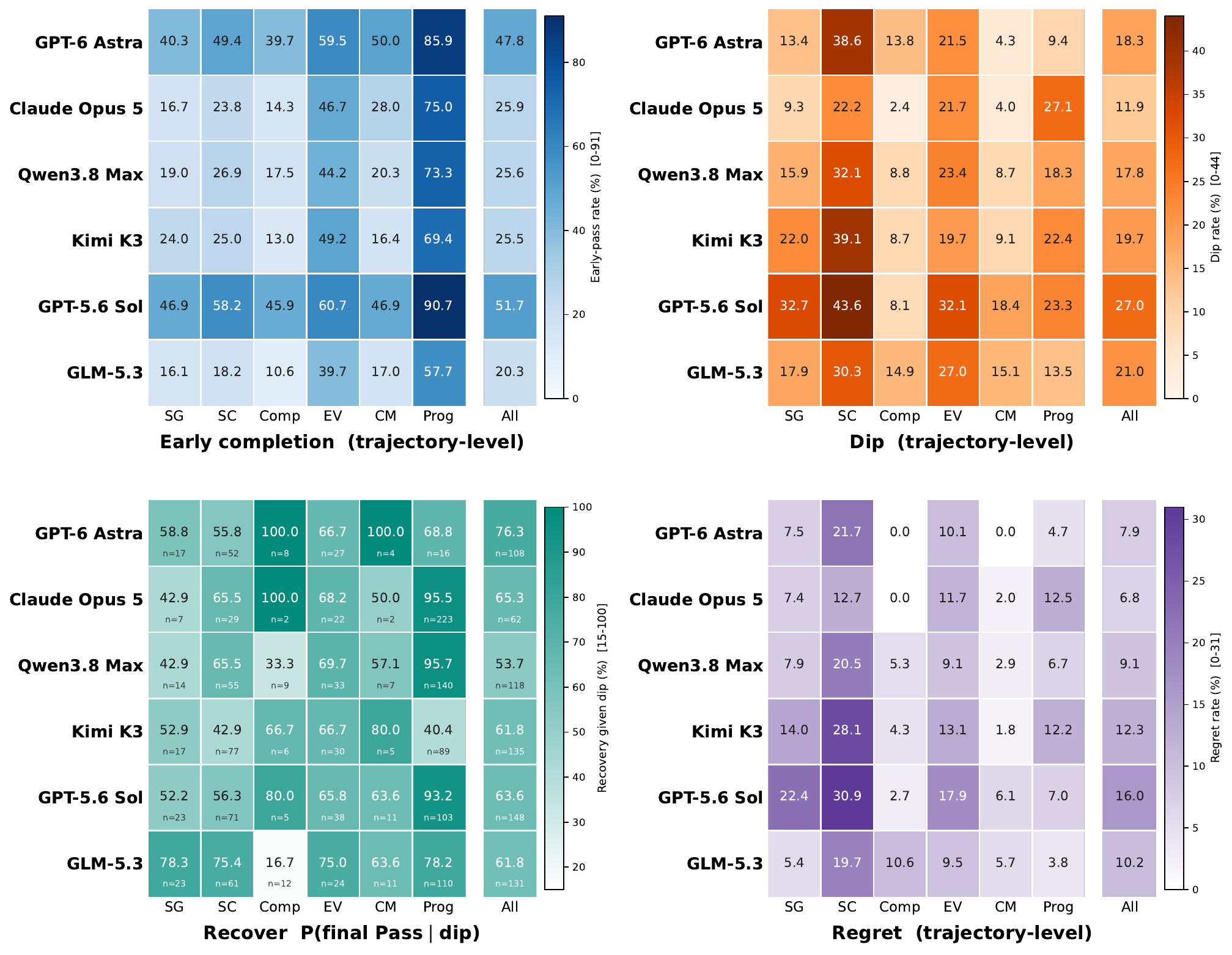}
    \caption{Behavior rates by model and evaluation dimension.}
    \label{fig:behaviors_heatmap}
\end{subfigure}
\caption{\textbf{Trajectory analysis of scientific agents.}
(a) Four examples illustrate four behaviors used in our analysis:
\emph{early completion} (reaching near-peak quality within the first third of snapshots),
\emph{dip} (a previously Pass becomes Fail),
\emph{recover} (a Failed rubric Passes again), and
\emph{regret} (a Pass rubric earlier but Fails at the final snapshot).
(b) Behavior rates across models four programmatic execution and five
LLM-judged dimensions of scientific report quality: Source Grounding,
Scientific Correctness, Completeness, Execution \& Validation, and
Communication.}
\label{fig:behaviors_res}
\end{figure*}

The results reveal a clear performance hierarchy, substantial room for
improvement, and distinct capability profiles across domains and task types.

\paragraph{\gptastra{} leads overall, but performance remains unsaturated.}
Table~\ref{tab:main} shows substantial variation across models.
\gptastra{} achieves the highest combined score of 80.9, outperforming
the second-ranked \claudefable{} by 13.6 points. Nevertheless, only
26.0\% of its runs achieve a normalized combined score of at least 0.95,
indicating considerable room for improvement.

\paragraph{Performance varies across domains and task types.}


The results reveal pronounced variation across domains and task types.
Figure~\ref{fig:heatmap_domains} shows that, although \gptastra{} leads
in 10 of the 11 domains, other models exhibit uneven capability profiles:
\claudefable{} scores 80 in Engineering but 49 in Bioinformatics,
while \kimik{} scores 81 in Materials Science but 50 in Physical Science.
Task-type results in Appendix~\ref{app:task_type_results} show that
\qwenmax{}, \glm{}, and \kimik{} perform relatively well on Literature
Synthesis.
These complementary capability profiles motivate evaluation along
both axes of SWTT.

\subsection{Hybrid Evaluation of Execution and Report Quality}

To capture both executable correctness and the quality of scientific
reporting, we adopt a hybrid evaluation framework that combines programmatic checks with rubric-based LLM judgment.

\textbf{Execution and report quality are only partially aligned.}
As shown in Figure~\ref{fig:prog_vs_judge}, the fraction of task-level cases
with strong reports but weak executable results (TL) ranges from 0.14 to 0.48
across models. \gptastra{} exhibits the strongest joint performance, with a
TR rate of 0.83 and a TL rate of only 0.14. In contrast, \deepseek{} has
the highest TL rate (0.48) and the lowest TR rate (0.25), showing that strong
report quality does not necessarily imply executable correctness.

\begin{figure}[t]
    \centering
    \includegraphics[width=0.9\linewidth]{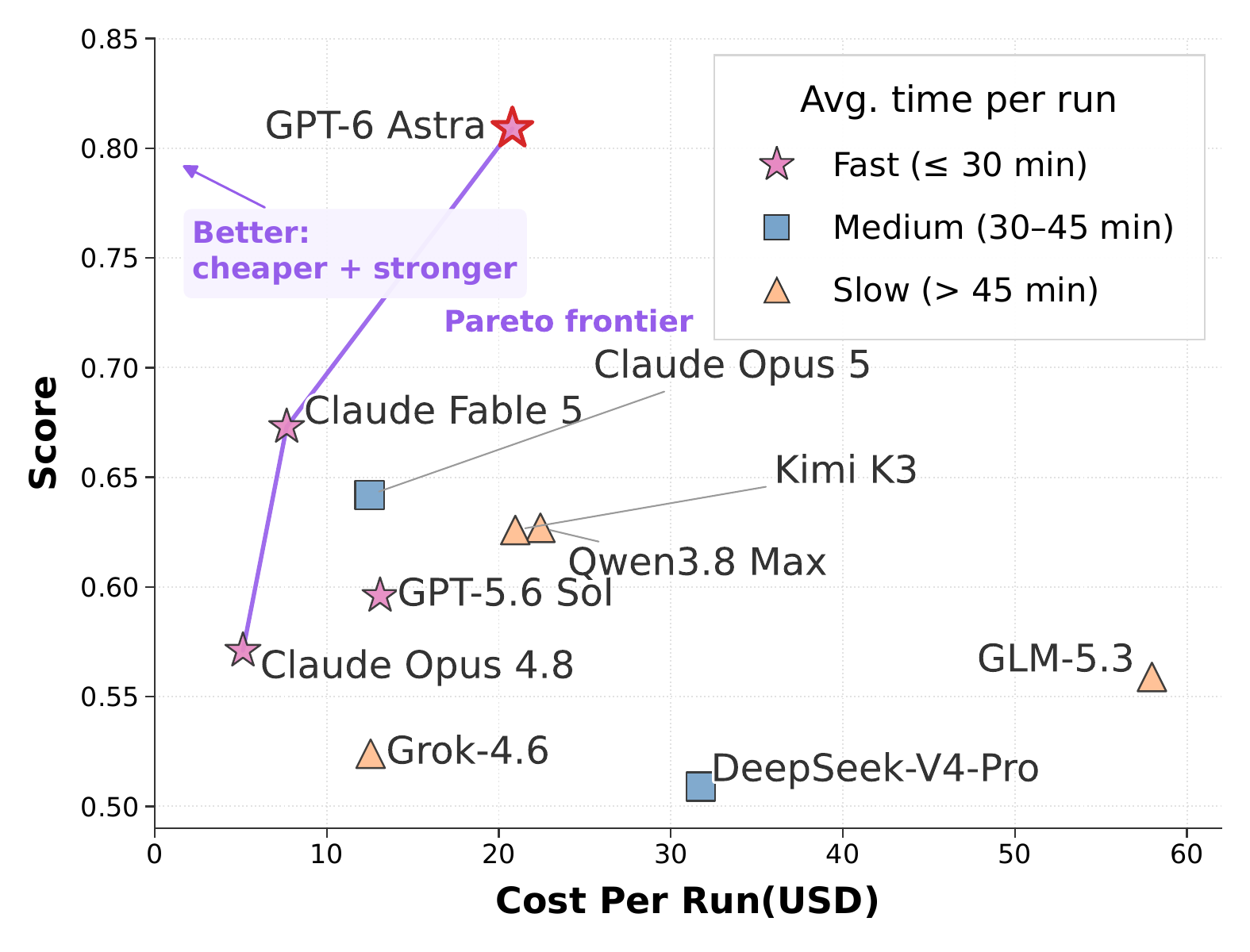}
    \caption{\textbf{Higher-priced models can be faster and cheaper per completed run.}
Overall score versus mean cost per run; the upper-left indicates higher performance at lower cost.}
    \label{fig:cost_overview}
    \vspace{-7pt}
\end{figure}

\textbf{Neither evaluation channel is sufficient on its own.}
Appendix~\ref{app:decoupling-case} illustrates two opposite failure modes.
\deepseek{} obtains a perfect programmatic score of 1.00 but only 0.43
from the LLM judge because its report omits key scientific interpretations.
Conversely, \claudeopusfive{} receives a perfect LLM-judge score of 1.00 but only
0.11 programmatically.
Evaluating either execution or reporting alone would therefore overlook a
critical aspect of task performance.

\subsection{Trajectory Behavior Analysis}

To analyze how solution stability evolves during execution, we replay trajectories across ten models, reconstruct a snapshot after each file read or write operation, and rescore each snapshot with both evaluation channels. We characterize four behaviors: \emph{early completion}, where near-peak quality is reached within the first third of the trajectory; \emph{dip}, where a previously passing criterion later fails; \emph{recover}, where a dipped criterion passes again; and \emph{regret}, where a previously passing criterion fails at the final snapshot.
Figure~\ref{fig:behaviors_res} reports behavior rates across models and evaluation dimensions for six models, with results for all ten models provided in Appendix~\ref{app:regret}.

\textbf{GPT models share an early-completion advantage but differ in solution stability.}
\gptastra{} and \gptsol{} exhibit the highest early-completion rates
(darker blue in Figure~\ref{fig:behaviors_res}b), indicating a shared tendency to reach
near-peak quality early in execution.
However, \gptsol{} has a higher mean dip rate across the five
rubric dimensions than \gptastra{} (27.0\% versus 18.3\%),
revealing greater susceptibility to subsequent regressions.
Trajectory analysis thus reveals both shared behavioral tendencies
and distinct weaknesses within the same model family.


\textbf{Programmatic regressions are generally more recoverable than scientific correctness failures.}
Programmatic checks exhibit higher recovery rates than Scientific
Correctness criteria in five of the six displayed models
(Figure~\ref{fig:behaviors_res}b).
\kimik{} is the exception, with recovery rates of 40.4\% and 42.9\%,
respectively. These results suggest that most evaluated agents restore
verifiable outcomes more reliably than the scientific validity of
their analyses.

\subsection{Efficiency and Cost Analysis}

To characterize agent efficiency, we measure mean API cost, runtime,
and interaction length per run.
Figure~\ref{fig:cost_overview} shows the cost--performance trade-off;
Appendix~\ref{app:cost} provides the full breakdown in
Table~\ref{tab:cost_vs_score} and Figure~\ref{fig:cost_bars}.


\textbf{Higher-priced models can achieve better results at lower total cost.}
Figure~\ref{fig:cost_overview} identifies \claudeopuseight{},
\claudefable{}, and \gptastra{} on the Pareto frontier,
offering competitive performance at different cost levels.
Despite its higher per-token price, \gptastra{} achieves the
highest combined score (80.9) at \$20.79 per run, outperforming
both \qwenmax{} and \kimik{} at lower total cost.
\glm{} incurs the highest cost (\$57.96), averaging
102.4 turns per run.
These results emphasize evaluating efficiency at the run level
rather than through token prices alone.

\section{Related Work}

\subsection{Science Benchmarks}

Scientific benchmarks assess a range of capabilities, but each
typically fixes on a single unit of evaluation. One line probes
scientific knowledge and reasoning through question
answering~\citep{wang2024scibench,sun2024scieval,laurent2024labbench},
data-driven discovery and computational
analysis~\citep{majumder2024discoverybench,mitchener2025bixbench},
or end-to-end discovery in interactive
environments~\citep{jansen2024discoveryworld}. Another centers on
executable artifacts: generated programs for data-driven
discovery~\citep{chen2025scienceagentbench}, computational
reproducibility and replication~\citep{siegel2024corebench,starace2025paperbench},
and research coding, code extension, and issue
resolution~\citep{yan2025lmrbench,hua2025researchcodebench,mmscicode2026,muller2025rexbench,swebenchscience2026}.
Each isolates a particular activity and a single output
modality, whether a textual answer or an executable result, leaving the
span of a full scientific workflow underassessed.
\sciswe{} instead organizes evaluation across seven scientific
workflow task types and requires both executable and textual
deliverables, measuring code-level correctness and scientific
judgment together.

\subsection{Scientific Agents}

ReAct and Toolformer integrate language-model reasoning with
external tools~\citep{yao2023react,schick2023toolformer}.
Building on this, autonomous scientific agents increasingly operate
across the research lifecycle: domain tool-using agents for chemistry
such as ChemCrow~\citep{mbran2024chemcrow} and the autonomous
experimentation system of~\citet{boiko2023coscientist}, general-purpose
biomedical agents such as Biomni~\citep{huang2025biomni}, multi-agent
discovery systems such as SciAgents~\citep{ghafarollahi2025sciagents},
and end-to-end frameworks that generate hypotheses, run experiments,
and draft papers such as The AI Scientist~\citep{lu2024aiscientist}
and AI-Researcher~\citep{tang2026ai}.
These systems emphasize pipeline autonomy and breadth, whereas
\sciswe{} targets standardized, stage-resolved evaluation of agent
behavior on well-scoped scientific workflow tasks.
MLE-bench and CSR-Bench evaluate multi-step execution in engineering
and research environments~\citep{chan2025mlebench,csrbench2025}.
Recent work studies memory management, specification following,
verification beyond functional tests, and adaptive
harnesses~\citep{swemem2026,specbench2026,swegate2026,openjiuwen2026}.
\sciswe{} examines scientific agents through both final output
quality and trajectory-level progress and stability.
\section{Conclusion}

We introduce \sciswe{}, a benchmark for evaluating language agents across the major stages of scientific inquiry. It includes 120 expert-curated tasks spanning 11 scientific domains and 7 task types of the Scientific Workflow Task Taxonomy, together with a hybrid evaluation protocol that assesses both computational execution and scientific reasoning.
We also find that stronger agents tend to make earlier progress during execution, while progress and solution stability capture distinct aspects of agent trajectories.
These findings highlight the importance of evaluating scientific agents beyond isolated task performance and provide insights for developing agents with broader capabilities and more effective and stable execution throughout scientific workflows.

\section*{Limitations}

While \textsc{Sci-Agent-Exam} provides a systematic evaluation of scientific agents in executable environments, the current study primarily focuses on general-purpose foundation models under standardized agent settings. As scientific agent systems continue to evolve, future work could extend the evaluation to specialized agents equipped with domain-specific tools, memory, planning, and iterative research capabilities. The benchmark could also be expanded with additional scientific workflows and evaluator configurations to support broader studies of scientific agents.

\bibliographystyle{plainnat}
\bibliography{iclr2027_conference}

\clearpage
\beginappendix
\section{AI Use Statement}

Generative AI tools were used during the preparation of this manuscript to
provide feedback on experimental analyses, improve the clarity and wording of
the text, and suggest potential titles and keywords. All AI-assisted revisions
and suggestions were critically reviewed and verified by the authors.
Generative AI tools were not used to generate benchmark data or experimental
results, develop the conceptual framework, or make final scientific decisions.

Separately, large language models are studied and employed as part of the
research methodology, including as evaluated agents and rubric-based
evaluators, as described in the main text. The authors take full responsibility
for the final text, claims, results, and artifacts presented in this work.

\section{Ethics Considerations}

The benchmark is constructed from publicly available scientific sources,
expert-authored tasks, and data used in accordance with their applicable
licenses or terms of use. It does not collect personal data or involve human
subjects as research participants. Some tasks concern biological or engineering
procedures, but they are designed as scientific analysis and verification
exercises rather than instructions for harmful activities.

All agent executions are conducted in isolated, offline environments with
restricted access to the host system, reducing the risks associated with
executing generated code. We document the provenance and applicable licenses
of released artifacts and discuss known limitations in domain representation.

Domain experts were recruited through a professional research collaboration
based on their relevant scientific expertise. They were informed that their
contributions would be used for benchmark construction and research. Their
participation and compensation followed applicable employment or contractual
arrangements.


\section{Benchmark Design and Construction}\label{app:task-package}
Each task is distributed as a versioned directory. The instruction states the scientific goal, constraints, expected files, and resource limits. The visible workspace contains only the files allowed for the task. Hidden inputs are mounted by the checker at evaluation time. The reference package contains a reference solution, expected schema, weighted checks, and a rubric. The metadata records domain, task type, capability tags, source provenance, license, and difficulty calibration. Every task that derives from external material documents, at the task level, the source it originates from together with that source's license. For example, a bioinformatics code-generation task reimplements the \emph{longdust} low-complexity-DNA algorithm introduced in a 2025 arXiv preprint (CC BY 4.0); a physics end-to-end analysis task is built on a particle-physics event-classification dataset released on Kaggle under CC0; and a physics tool-orchestration task builds on the open-source Smilei particle-in-cell simulation code. We include only sources whose license permits redistribution and derivative use, and licenses are respected throughout the benchmark.

\subsection{Task and domain distributions}
\label{app:detailed_distribution}

As shown in Table~\ref{tab:domain_type_distribution}, \sciswe{} achieves broad
and well-distributed coverage across scientific domains and workflow types.
Most domains span multiple task types, avoiding strong concentration in any
single domain--type combination.


\paragraph{Human-authored rubrics.}
The evaluation rubrics are authored by the same domain experts who construct
the tasks and are not generated by an LLM. Rubrics are written after the
reference solution is completed, allowing their criteria to reflect the
requirements of an expert-quality solution. As a calibration check, the
reference solution's report must achieve above 90\% under its corresponding
rubric before the task is released. LLMs are introduced only at evaluation
time, where the LLM Judge applies these human-written criteria to agent
outputs rather than generating or modifying the rubric itself.

\subsection{Expert construction and verification}

A total of 156 domain experts participated in task construction and
verification, with data-production and evaluation personnel kept independent.
Each task proceeds through three roles. A task writer selects the source,
designs and implements the task, constructs the reference solution, and
defines the evaluation criteria. An independent reviewer then checks the
scientific validity, task specification, reference workflow, executable
checks, and rubric. A separate auditor verifies the review decision and
checks consistency with the construction guidelines.

Writers targeted approximately six hours per task across task design,
implementation, and rubric construction, while reviewers and auditors
targeted approximately two hours and one hour per task, respectively.
Reviewer--auditor agreement was subject to an 85\% quality threshold;
cases below this threshold were returned for further review or calibration.

\subsection{Representative tasks}
Representative instructions include a bioinformatics survival analysis that tests four conclusions under interval and perturbation choices. A protein conformation workflow must recover a decisive formula from a paper, compute an evidence surface, and perform leave one measurement and energy population sensitivity checks. A code task implements three compositional analysis functions with a self test and a reproducible answer file. A result interpretation task reconciles conflicting tetM selection claims by comparing data range, methods, and functional evidence. A hypothesis task requires a design artifact before observing results and separates model fit from causal support. An error diagnosis task repairs a silent regression that changes the scientific sample while leaving tests green. These examples illustrate why scientific-agent evaluation must assess both
successful execution and the scientific presentation of the resulting
evidence and conclusions.

\subsection{Details of Trap Injection}
\label{app:trap}

\definecolor{trapshade}{RGB}{250,238,238}   
\definecolor{solveshade}{RGB}{238,247,240}  

Many \sciswe{} tasks are built around a \emph{trap}: the obvious method, the loudest signal,
or a ready-made result is engineered to be wrong, and credit requires seeing past it.
Figure~\ref{fig:trap-card} shows one task verbatim, contrasting the two paths an agent can take from the same workspace following the decoy versus rejecting it.

\begin{figure}[t]
\centering
\begingroup
\small
\setlength{\fboxsep}{0pt}
\setlength{\fboxrule}{0.4pt}

\fbox{%
\begin{minipage}{\dimexpr\linewidth-2\fboxrule\relax}

\begingroup
\setlength{\fboxsep}{6pt}
\colorbox{headergray}{%
\parbox{\dimexpr\linewidth-2\fboxsep\relax}{%
\raggedright
\textbf{Trap card | Hypothesis Validation}\\[4pt]
{\footnotesize
Programmatic: 50\% \quad LLM rubric: 50\%}\\[4pt]
{\scriptsize
\texttt{data-science-hypothesis-validation-01}}\\
{\footnotesize Data Science}
}}
\endgroup

\vspace{6pt}

\noindent\hspace*{6pt}%
\begin{minipage}{\dimexpr\linewidth-12pt\relax}
\raggedright
\textbf{Task.}
Validate whether the \texttt{DRUG\_AZ\_17} biomarker screen
is suitable for advancement under the mounted article's
cross-validation and preprocessing construction.

\medskip
\textbf{The decoy.}
The workspace includes a \texttt{screening\_release\_v2}
rollup with validation metrics, gate booleans, a recommended
action, and a verdict, formatted like a finished answer.
\end{minipage}

\vspace{8pt}

\noindent\hspace*{6pt}%
\begin{minipage}{\dimexpr\linewidth-12pt\relax}
\setlength{\fboxsep}{5pt}
\noindent
\begin{minipage}[t]{\dimexpr0.5\linewidth-3pt\relax}
\vspace{0pt}%
\colorbox{trapshade}{%
\begin{minipage}[t]{\dimexpr\linewidth-2\fboxsep\relax}
\raggedright
\footnotesize
\textbf{Follow the decoy}
{\color{red!70!black}\ding{55}}

\medskip
\textit{Action.}
Copy the rollup's verdict and action, or screen features
on the full matrix before cross-validation.

\medskip
\textit{Answer.}\\
\texttt{supported}\\
\texttt{advance\_screen}

\medskip
\textit{Verification.}
The checker recomputes the gates and reruns
\texttt{run\_analysis.py} on modified data and fold
assignments. Copied or hard-coded decisions fail
these checks.
\end{minipage}}
\end{minipage}%
\hfill
\begin{minipage}[t]{\dimexpr0.5\linewidth-3pt\relax}
\vspace{0pt}%
\colorbox{solveshade}{%
\begin{minipage}[t]{\dimexpr\linewidth-2\fboxsep\relax}
\raggedright
\footnotesize
\textbf{Reject the decoy}
{\color{green!45!black}\ding{51}}

\medskip
\textit{Action.}
Reconstruct validation from the article. Fit screening
and scaling on each fold's training split; leave binary
\texttt{*.mut.} columns unscaled and remove duplicate
feature signatures.

\medskip
\textit{Answer.}\\
\texttt{not\_supported}\\
\texttt{hold\_for\_redesign}

\medskip
\textit{Verification.}
All three gates fail on the original inputs.
The submitted analysis is rerun and checked under
input mutation.
\end{minipage}}
\end{minipage}
\end{minipage}

\vspace{6pt}
\end{minipage}%
}
\endgroup

\caption{\textbf{Example of an adversarial trap.}
A ready-made rollup computed with data leakage leads to
an incorrect decision. Correct validation fits preprocessing
within each training fold before evaluating held-out data.
Input-mutation checks test whether the submitted code
recomputes the result rather than copying a fixed verdict.
Such traps are documented in approximately 85\% of tasks;
input-mutation checks are used in 61\%.}
\label{fig:trap-card}
\end{figure}


The same principle recurs in a few forms across the benchmark
(Table~\ref{tab:trap-forms}). They share a common structure: a locally reasonable shortcut
is placed within reach, and only the full scientific procedure separates it from the
reference answer. Where the scientific contract admits it, the shortcut is defeated
mechanically by input mutation; otherwise it is defeated by withheld references and hidden
reruns (Appendix~\ref{app:task-package}).

\begin{table*}[t]
\centering
\small
\setlength{\tabcolsep}{5pt}
\renewcommand{\arraystretch}{1.3}
\begin{tabular}{@{}>{\raggedright\arraybackslash}p{2.9cm}>{\raggedright\arraybackslash}p{6.4cm}>{\raggedright\arraybackslash}p{3.1cm}@{}}
\toprule
\textbf{Form} & \textbf{The decoy $\rightarrow$ what credit requires} & \textbf{Representative task} \\
\midrule
Decoy artifact &
A precomputed rollup or dashboard reads like the answer but was produced by a leaky or
partial procedure $\rightarrow$ recompute the decision independently. &
\texttt{data-science-\allowbreak hypothesis-\allowbreak validation-01} \\
Loudest signal &
The most salient statistic (lowest information criterion, lowest cost) nominates a
candidate that fails a downstream physical or feasibility gate $\rightarrow$ select on the
gated objective. &
\texttt{engineering-\allowbreak tool-\allowbreak orchestration-02} \\
Distractor feature &
Balanced noise covariates masquerade as signal $\rightarrow$ keep only the load-bearing
variables. &
\texttt{bioinformatics-\allowbreak end-to-end-\allowbreak analysis-05} \\
Misattributed cause &
A dashboard blames the wrong component and discourages the real fix $\rightarrow$ diagnose
and repair the true failure path. &
\texttt{cs-expansion-\allowbreak error-\allowbreak diagnosis-02} \\
\bottomrule
\end{tabular}
\caption{Recurring trap forms in \sciswe{}, with a representative task for each. In every
form the decoy is locally plausible; credit requires the coupled scientific workflow.}
\label{tab:trap-forms}
\end{table*}

\subsection{Task type weights}
\label{app:task_type_weights}

The task-level score combines programmatic and LLM-judge scores using
task-type-specific weights fixed before model evaluation
(Table~\ref{tab:task_type_weights}). We report both component scores
alongside the aggregate to make performance on each evaluation component
visible.

\begin{table}[t]
\centering
\small
\setlength{\tabcolsep}{7pt}
\begin{tabular}{lcc}
\toprule
\textbf{Task type} & $\boldsymbol{w_i^p}$ & $\boldsymbol{w_i^j}$ \\
\midrule
Code Generation       & 0.70 & 0.30 \\
Error Diagnosis       & 0.65 & 0.35 \\
Tool Orchestration    & 0.55 & 0.45 \\
End-to-End Analysis   & 0.50 & 0.50 \\
Hypothesis Validation & 0.50 & 0.50 \\
Result Interpretation & 0.45 & 0.55 \\
Literature Synthesis  & 0.20 & 0.80 \\
\bottomrule
\end{tabular}
\caption{Programmatic and LLM-judge weights by task type.}
\label{tab:task_type_weights}
\end{table}

For a model $m$ and task type $k$, we compute the macro average
\begin{equation}
 S_{m,k}=\frac{1}{|T_k|}\sum_{i\in T_k}s_i.
\end{equation}
Domain averages use the same task level scores and do not reweight domains.






\section{Evaluation Robustness and Sensitivity}
\label{app:evaluation_robustness}

We examine sensitivity to the agent harness and inference-effort setting
on a 50-task subset. These analyses complement the main evaluation;
scores under different harnesses are averaged over their respective
clean-task sets, whereas the effort comparison uses paired tasks.

\subsection{Robustness across Agent Harnesses}
\begin{table*}[t]
\centering
\small
\begin{tabular}{llcccc}
\toprule
Model & Harness & $n$ & Prog. & Judge & Final \\
\midrule
\multirow{2}{*}{gpt-6-astra}
  & claude-code   & 42 & 72.3 & 85.8 & 78.9 \\
  & codex             & 45 & 74.8 & 83.8 & \textbf{79.5} \\
\midrule
\multirow{3}{*}{qwen-3.8-max}
  & claude-code   & 49 & 65.6 & 75.9 & 70.0 \\
  & codex             & 46 & 67.4 & 76.6 & \textbf{71.3} \\
  & qoder             & 40 & 62.9 & 73.5 & 67.0 \\
\midrule
\multirow{3}{*}{kimi-k3}
  & claude-code   & 47 & 65.8 & 75.3 & \textbf{69.9} \\
  & codex             & 47 & 50.1 & 59.3 & 54.0 \\
  & kimi-cli          & 49 & 54.5 & 63.4 & 58.5 \\
\midrule
\multirow{2}{*}{glm-5.3}
  & claude-code   & 49 & 60.7 & 74.1 & \textbf{66.2} \\
  & codex             & 47 & 43.8 & 49.4 & 46.1 \\
\midrule
\multirow{2}{*}{claude-opus-4-8}
  & claude-code   & 49 & 59.3 & 66.7 & \textbf{62.2} \\
  & codex             & 47 & 57.0 & 55.8 & 55.6 \\
\midrule
\multirow{2}{*}{gpt-5.6-sol}
  & claude-code   & 50 & 58.4 & 52.1 & 55.6 \\
  & codex             & 45 & 72.9 & 66.8 & \textbf{69.9} \\
\bottomrule
\end{tabular}
\caption{Harness sensitivity on the 50-task set: the same model scored under
general-purpose harnesses (claude-code at max effort, codex) and, where available,
its vendor-native harness (kimi-cli for kimi-k3, qoder for qwen-3.8-max).}
\label{tab:multi-harness}
\end{table*}

Table~\ref{tab:multi-harness} shows that harness choice can materially
affect measured performance, although the effect varies by model.
\gptastra{} obtains similar Final scores under claude-code and codex
(78.9 versus 79.5), as does \qwenmax{} (70.0 versus 71.3).
In contrast, \kimik{} and \glm{} score 15.9 and 20.1 points lower,
respectively, under codex than under claude-code. \gptsol{} shows
the opposite pattern, scoring 14.3 points higher under codex.
These comparisons indicate that benchmark scores should be interpreted
together with the harness configuration. Because each harness row
averages over its own clean tasks, the differences are descriptive
rather than paired estimates of a harness effect.

\subsection{Sensitivity to Inference Effort}
\begin{table*}[t]
\centering
\small
\begin{tabular}{llcccc}
\toprule
Model & Effort & Programmatic & LLM Judge & Final & Pass@0.95 \\
\midrule
\multirow{2}{*}{gpt-6-astra ($n{=}42$)}
  & high & 74.5 & 83.8 & 78.9 & 5\% \\
  & max  & 72.3 & 85.8 & 78.9 & 21\% \\
\midrule
\multirow{2}{*}{qwen-3.8-max ($n{=}49$)}
  & high & 59.1 & 73.6 & 65.4 & 0\% \\
  & max  & 65.6 & 75.9 & 70.0 & 4\% \\
\midrule
\multirow{2}{*}{kimi-k3 ($n{=}47$)}
  & high & 61.2 & 70.3 & 64.7 & 0\% \\
  & max  & 65.8 & 75.3 & 69.9 & 6\% \\
\midrule
\multirow{2}{*}{claude-fable-5 ($n{=}38$)}
  & high & 56.2 & 64.1 & 59.3 & 3\% \\
  & max  & 64.9 & 69.8 & 66.6 & 5\% \\
\midrule
\multirow{2}{*}{glm-5.3 ($n{=}49$)}
  & high & 59.9 & 74.4 & 63.1 & 2\% \\
  & max  & 60.7 & 74.1 & 66.2 & 4\% \\
\midrule
\multirow{2}{*}{claude-opus-4-8 ($n{=}49$)}
  & high & 55.2 & 59.1 & 56.0 & 0\% \\
  & max  & 59.3 & 66.7 & 62.2 & 4\% \\
\midrule
\multirow{2}{*}{grok-4.6 ($n{=}45$)}
  & high & 59.3 & 67.0 & 60.0 & 0\% \\
  & max  & 61.9 & 65.8 & 61.1 & 4\% \\
\midrule
\multirow{2}{*}{deepseek-v4-pro-0813 ($n{=}49$)}
  & high & 45.2 & 58.5 & 50.8 & 0\% \\
  & max  & 54.2 & 59.4 & 55.8 & 2\% \\
\midrule
\multirow{2}{*}{gpt-5.6-sol ($n{=}50$)}
  & high & 58.2 & 58.8 & 58.0 & 0\% \\
  & max  & 58.4 & 52.1 & 55.6 & 2\% \\
\midrule
\multirow{2}{*}{\textbf{Overall (9 models)}}
  & high & 58.7 & 67.7 & 61.8 & 1\% \\
  & max  & 62.6 & 69.4 & 65.1 & 6\% \\
\bottomrule
\end{tabular}
\caption{Effect of \emph{max} vs.\ \emph{high} thinking effort on the claude-code
harness (paired per task on the 50-task set; $n$ = paired tasks). Programmatic, LLM
Judge and Final are on a $0$--$100$ scale; Final is the per-task mean reward averaged
over tasks; Pass@0.95 is the fraction of tasks with mean final reward $\geq 95$. Rows
ordered by max-effort Final.}
\label{tab:effort-max-vs-high}
\end{table*}

Table~\ref{tab:effort-max-vs-high} compares high and max effort on
paired tasks using the same claude-code harness. Averaged across the
nine models, Final increases from 61.8 to 65.1, and Pass@0.95 rises
from 1\% to 6\%. The gains are uneven: \claudefable{}, \claudeopuseight{},
\kimik{}, and \deepseek{} gain 5.0--7.3 Final points, while
\gptastra{} remains at 78.9 and \gptsol{} decreases from 58.0
to 55.6. Thus, additional inference effort improves aggregate
performance on this subset but does not reliably improve every model
or every evaluation component.
\section{Experimental Details and Cost}
\label{app:cost}

\paragraph{Orchestration framework.}
All evaluations are executed with the open-source \texttt{harbor} runner
\citep{harbor_2026} (v0.20.0), a terminal-bench-style harness. For each task it
builds the task's Docker image, invokes the agent in the container, collects the
produced files, runs the deterministic checker and the rubric-based judge, and
writes the aggregated scalar reward.

\begin{table*}[t]
\centering
\small
\setlength{\tabcolsep}{4pt}
\renewcommand{\arraystretch}{1.08}

\begin{tabular}{@{}lcccc@{}}
\toprule
\textbf{Model} &
\textbf{Scores} &
\textbf{Avg. Cost/Run (\$)} &
\textbf{Runtime (min)} &
\textbf{Turns} \\
\midrule
\gptastra{} & \textbf{80.9} & 20.79 & 22.1 & 56.8 \\
\claudefable{}        & 67.3 &  7.67 & 25.3 & 30.2 \\
\claudeopusfive{}         & 64.2 & 12.49 & 30.6 & 43.9 \\
\qwenmax{}           & 62.7 & 22.42 & 50.9 & 59.6 \\
\kimik{}               & 62.6 & 20.96 & 47.2 & 52.6 \\
\gptsol{}           & 59.6 & 13.10 & \textbf{22.0} & 54.3 \\
\claudeopuseight{}       & 57.1 & \textbf{5.13} & 23.3 & 37.3 \\
\glm{}               & 55.9 & 57.96 & 56.6 & 102.4 \\
\grok{}              & 52.4 & 12.55 & 45.8 & 44.0 \\
\deepseek{}       & 50.9 & 31.73 & 30.1 & 63.5 \\
\bottomrule
\end{tabular}

\vspace{2pt}

\caption{
Performance, average cost, runtime, and interaction length on the \sciswe{} peer subset.
}
\label{tab:cost_vs_score}

\end{table*}

\begin{figure}[t]
    \centering
    \includegraphics[width=\linewidth]{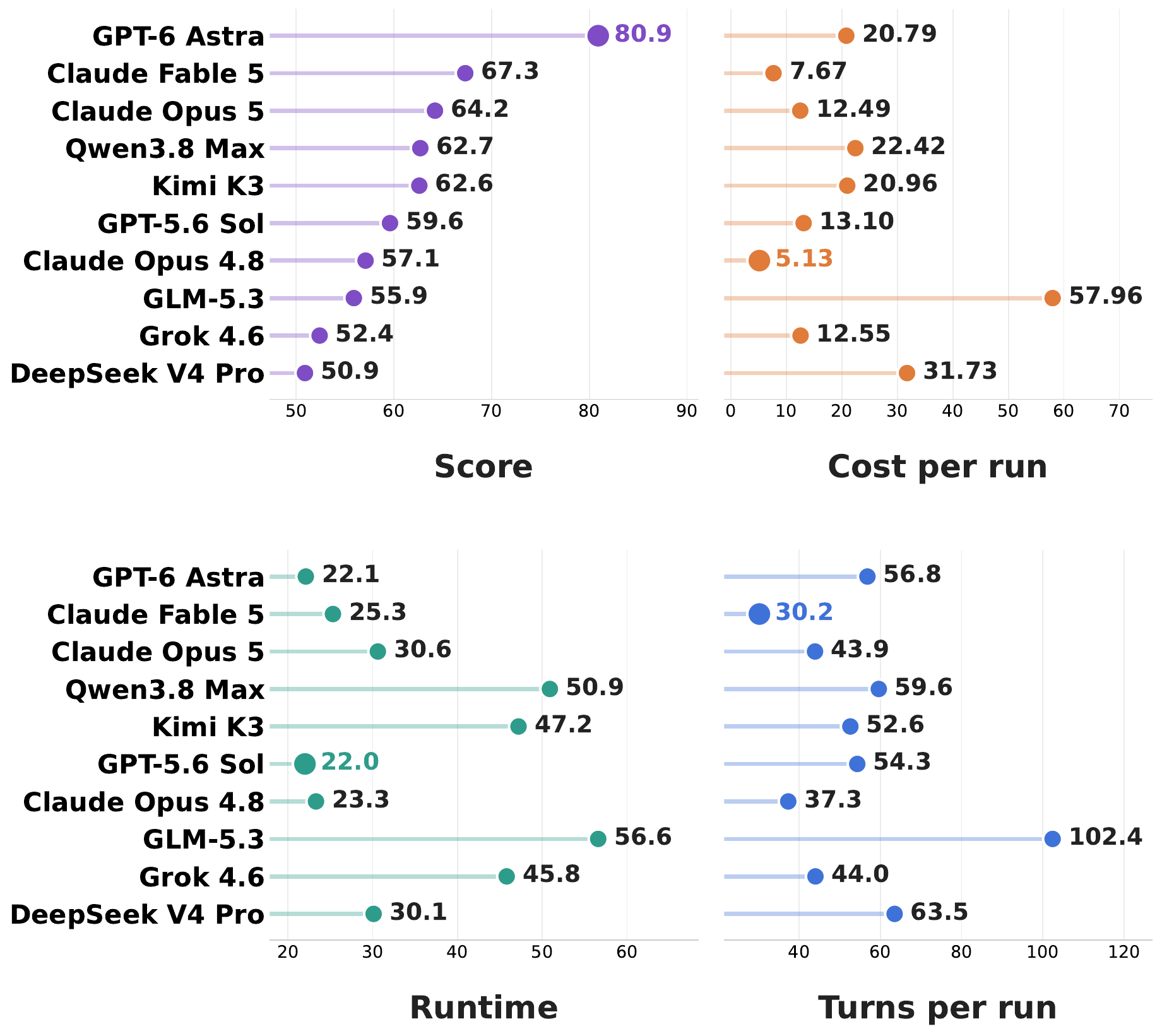}
    \caption{\textbf{Performance and resource use across ten agents on \sciswe{}.}
    Overall score, mean cost, runtime, and turns per run, ordered by overall score. The best value for each metric is highlighted.}
    \label{fig:cost_bars}
\end{figure}

Table~\ref{tab:cost_vs_score} and Figure~\ref{fig:cost_bars} report model-level performance and
per-run resource use on the peer subset. Cost denotes the observed
mean expenditure for a complete agent run, rather than a provider's
unit token price; runtime and turns describe the time and interaction
required under the same harness setting.

The results show that lower run cost does not necessarily coincide
with stronger task performance. \gptastra{} obtains the highest
combined score (80.9) at \$20.79 per run and 22.1 minutes on average.
By comparison, \qwenmax{} and \kimik{} cost \$22.42 and \$20.96
per run, respectively, while scoring 62.7 and 62.6 and taking
substantially longer. \claudefable{} presents a different trade-off:
it reaches the second-highest score (67.3) at the lowest cost among
the top five models (\$7.67 per run). At the other end of the cost
range, \glm{} averages \$57.96 and 102.4 turns per run, yet scores
55.9. These comparisons illustrate why unit token prices alone
cannot characterize the cost of an agentic scientific workflow:
the number of interactions, their duration, and the resulting task
quality all matter.

Turns offer a partial view of interaction efficiency, but should
not be interpreted as a direct measure of reasoning quality or
wasted steps. Models can differ in tokens per turn, tool use, and
provider pricing, and the table does not isolate these factors.
Accordingly, we report observed per-run cost together with score,
runtime, and turns, allowing readers to assess the performance and
resource trade-offs under this specific harness configuration.

\section{Results across Task Types and Rubric Dimensions}
\label{app:task_type_results}

We provide two complementary decompositions of model performance beyond the domain-level analysis in the main text. We first examine performance across the seven task types in \sciswe{}, which characterize different stages of scientific work, and then decompose the rubric-based evaluation into five dimensions of scientific quality.

\begin{figure*}[t]
\centering
\includegraphics[width=0.99\linewidth]{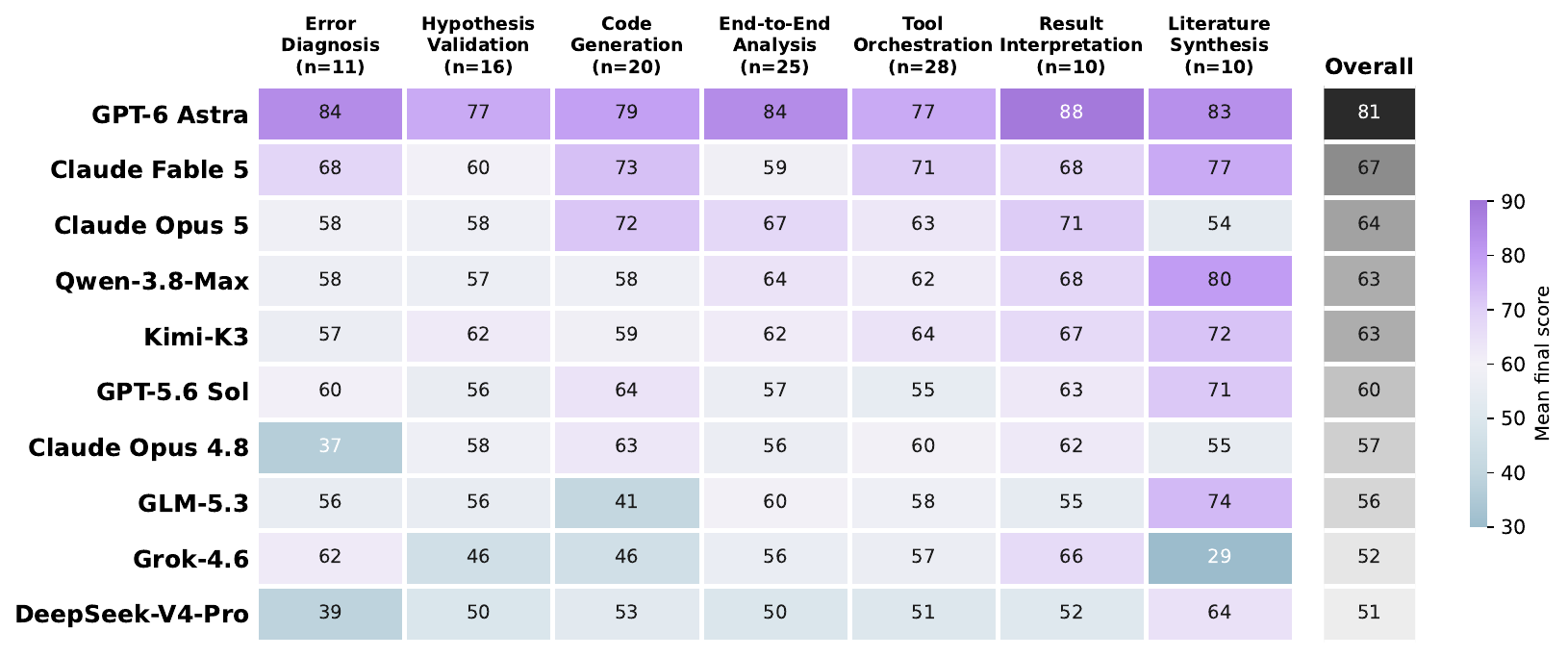}
\caption{
\textbf{Model performance across the seven task types in \sciswe{}.}
Each cell reports the mean final task score for a model--task-type pair,
with the number of tasks shown in parentheses.
Rows are ordered by overall benchmark performance, and the rightmost
column reports each model's overall final score.
}
\label{fig:heatmap_tasks}
\end{figure*}

\paragraph{Performance across task types.}
Figure~\ref{fig:heatmap_tasks} reveals distinct capability profiles across
scientific activities. \gptastra{} is consistently strong, leading all seven
task types with scores between 77 and 88. The remaining models are considerably
less uniform. Notably, models with similar overall scores can succeed on
different types of scientific work: \claudefable{} and \claudeopusfive{} are
stronger on Code Generation and Tool Orchestration, whereas \qwenmax{},
\kimik{}, and \glm{} perform relatively well on Literature Synthesis.
The contrast is particularly pronounced for \glm{}, which scores 74 on
Literature Synthesis but only 41 on Code Generation, and for \grok{},
whose scores range from 29 on Literature Synthesis to 66 on Result
Interpretation.

Task types also differ in how strongly they separate models. Literature
Synthesis exhibits the widest range (29--83), whereas several other task
types produce more compressed performance differences. Overall, the results
show that aggregate performance does not correspond to a single transferable
notion of scientific capability: models with comparable final scores can have
substantially different strengths across research activities.

\begin{figure}[t]
\centering
\includegraphics[width=0.9\linewidth]{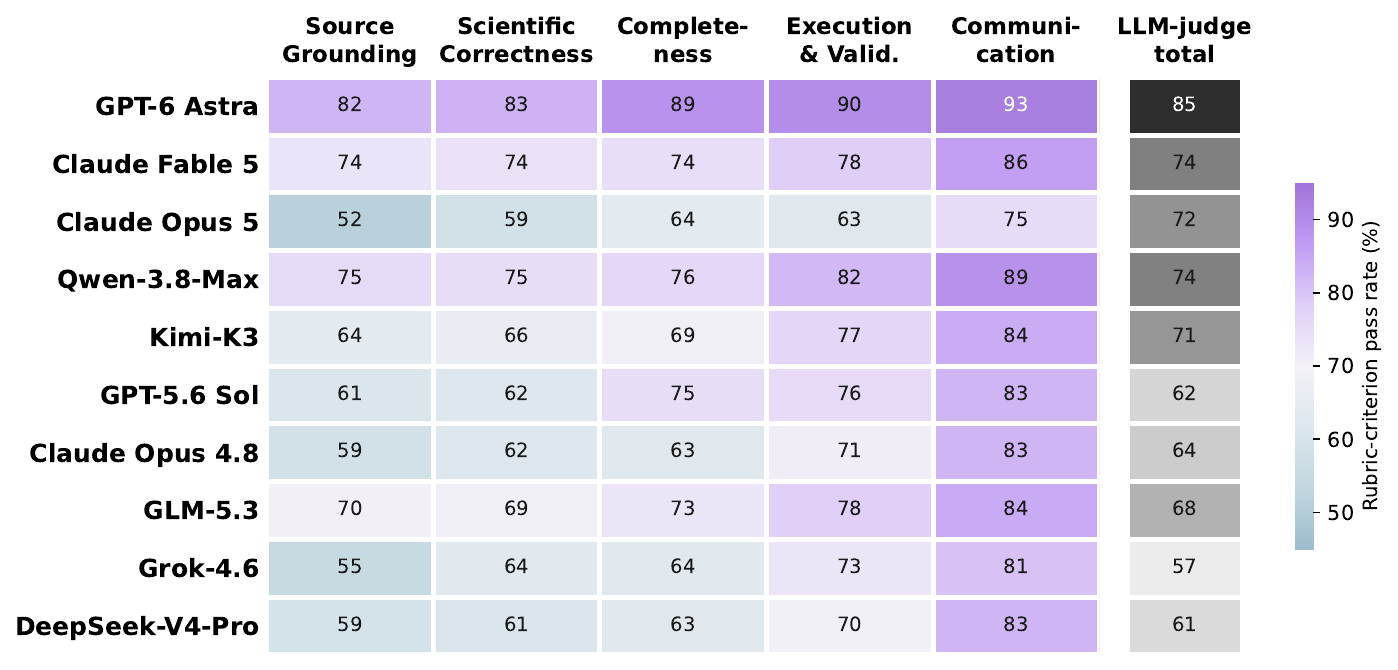}
\caption{
\textbf{Rubric-based performance across five dimensions of scientific quality.}
Each cell reports the fraction of rubric criteria passed within a dimension for each model.
The five dimensions capture Source Grounding, Scientific Correctness,
Completeness, Execution \& Validation, and Communication.
The rightmost column reports the aggregate LLM-judge score.
}
\label{fig:heatmap_rubric}
\end{figure}

\paragraph{Performance across rubric dimensions.}
Figure~\ref{fig:heatmap_rubric} reveals a different but highly consistent
pattern across models: agents are generally stronger at communicating and
presenting their results than at grounding and establishing their scientific
correctness. Communication is the highest-scoring dimension for every model,
reaching 81--93\% criterion pass rates, while Source Grounding and Scientific
Correctness are typically among the weakest. This gap persists even for the
strongest model: \gptastra{} reaches 93\% on Communication and 90\% on
Execution \& Validation, compared with 82\% on Source Grounding and 83\% on
Scientific Correctness.

The decomposition also distinguishes models with similar aggregate judge
scores. \claudefable{} and \qwenmax{} both score 74 overall, yet \qwenmax{}
is stronger on Execution \& Validation (82 vs.\ 78) and Communication
(89 vs.\ 86), while their remaining dimensions are similar. More broadly,
high presentation quality does not imply equally reliable scientific
grounding or correctness. The aggregate LLM-judge score therefore compresses
an important distinction between \emph{how well a scientific result is
presented} and \emph{how well it is scientifically supported}.

\section{Trajectory-Level Behavior Patterns}
\label{app:regret}

\begin{figure}[t]
    \centering
    \begin{subfigure}[b]{\linewidth}
        \centering
        \includegraphics[width=\linewidth]{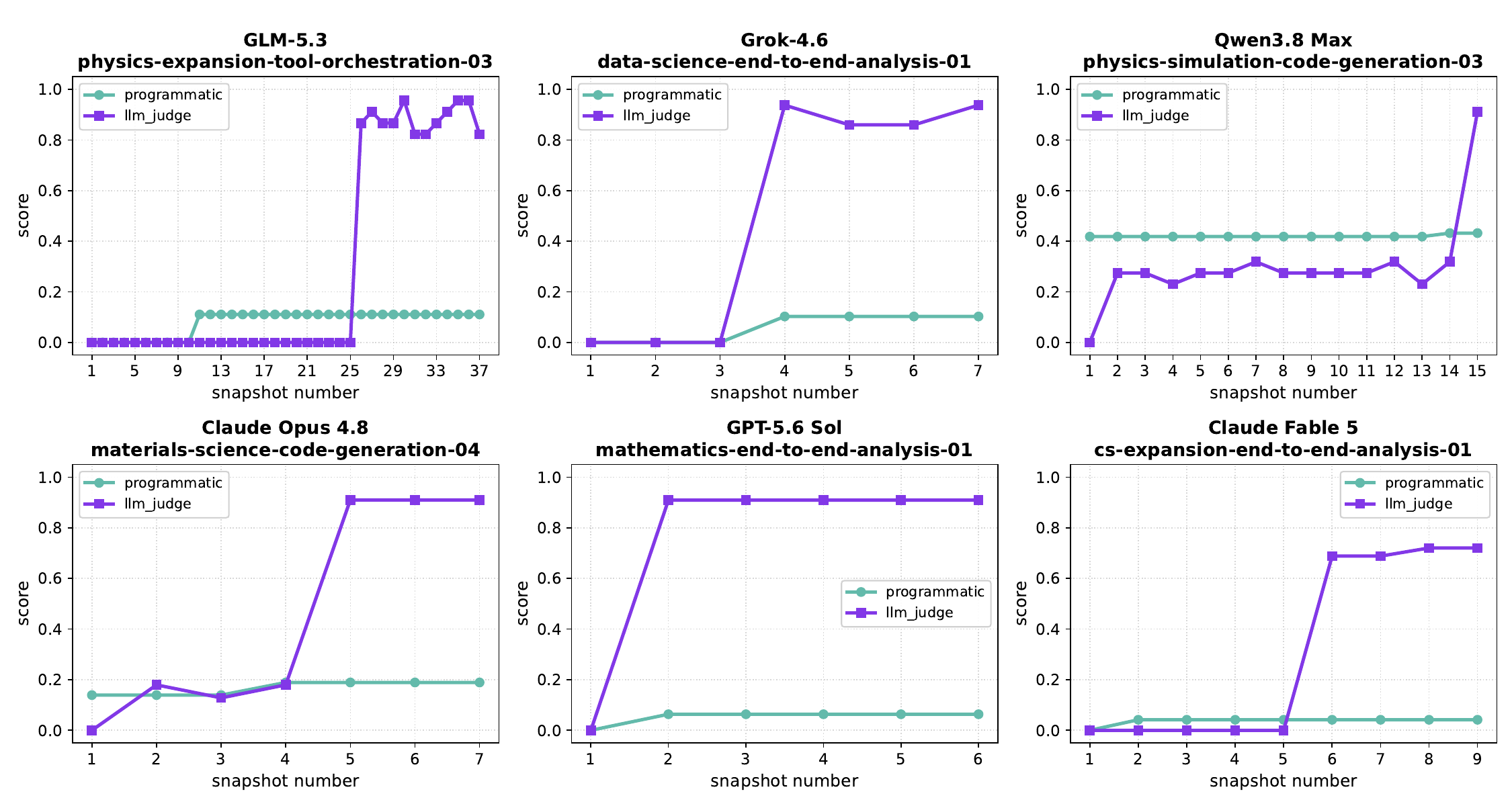}
        \caption{Typical trajectory patterns across different models and tasks.}
        \label{fig:exemplar_typical}
    \end{subfigure}
    
    \vspace{1em} 
    
    \begin{subfigure}[b]{\linewidth} 
        \centering
        \includegraphics[width=\linewidth]{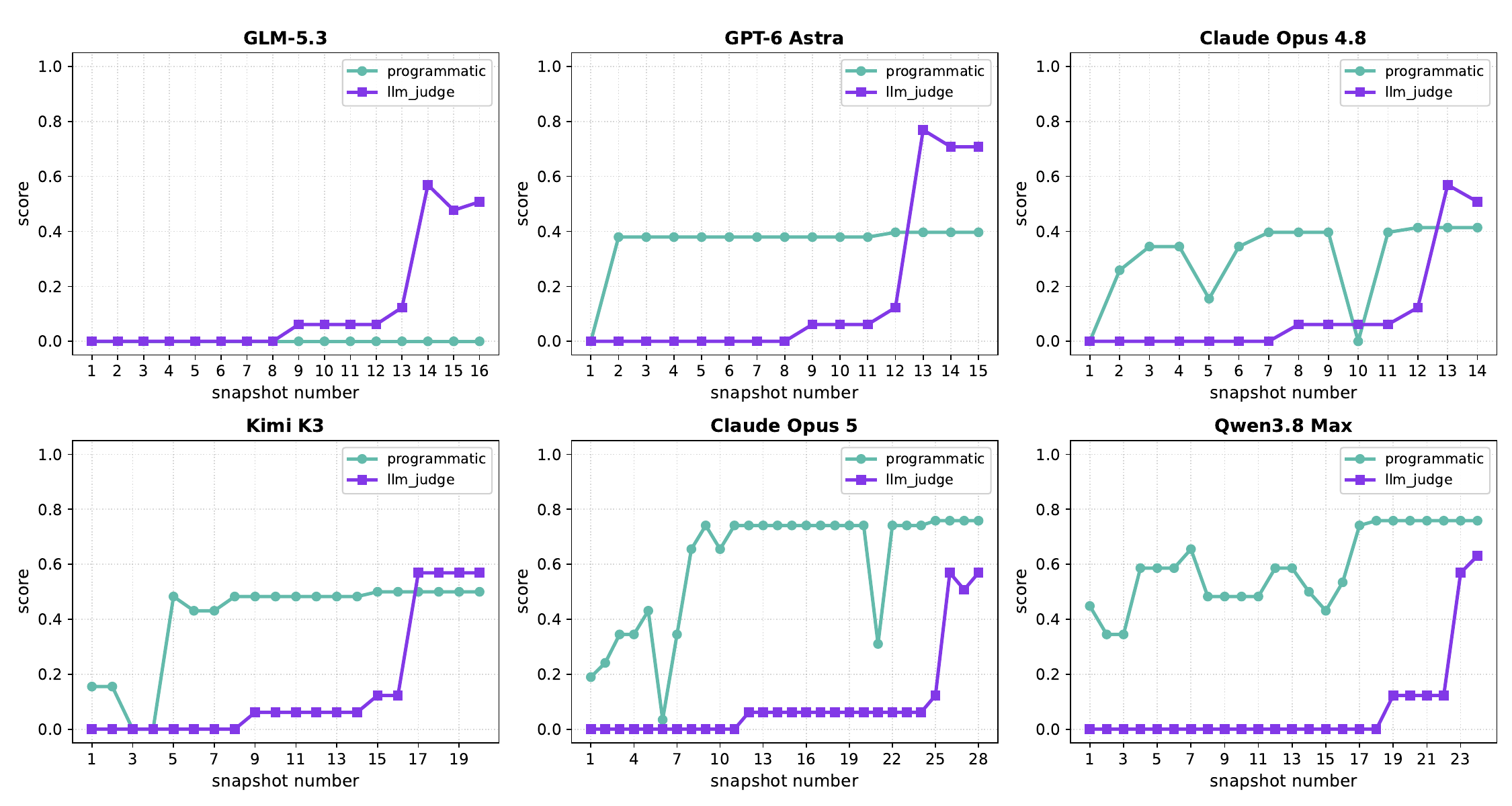}
        \caption{Model performance variations on the same task.}
        \label{fig:exemplar_same_task}
    \end{subfigure}
    
    \caption{Trajectory-level analysis of agent behaviors. \textbf{(a)} illustrates the diverse score dynamics (e.g., oscillations, late-stage jumps) across different models. \textbf{(b)} highlights the performance variations when multiple models tackle the same scientific task.}
    \label{fig:exemplar_combined}
\end{figure}

\begin{figure*}[t]
    \centering
    \includegraphics[width=\linewidth]{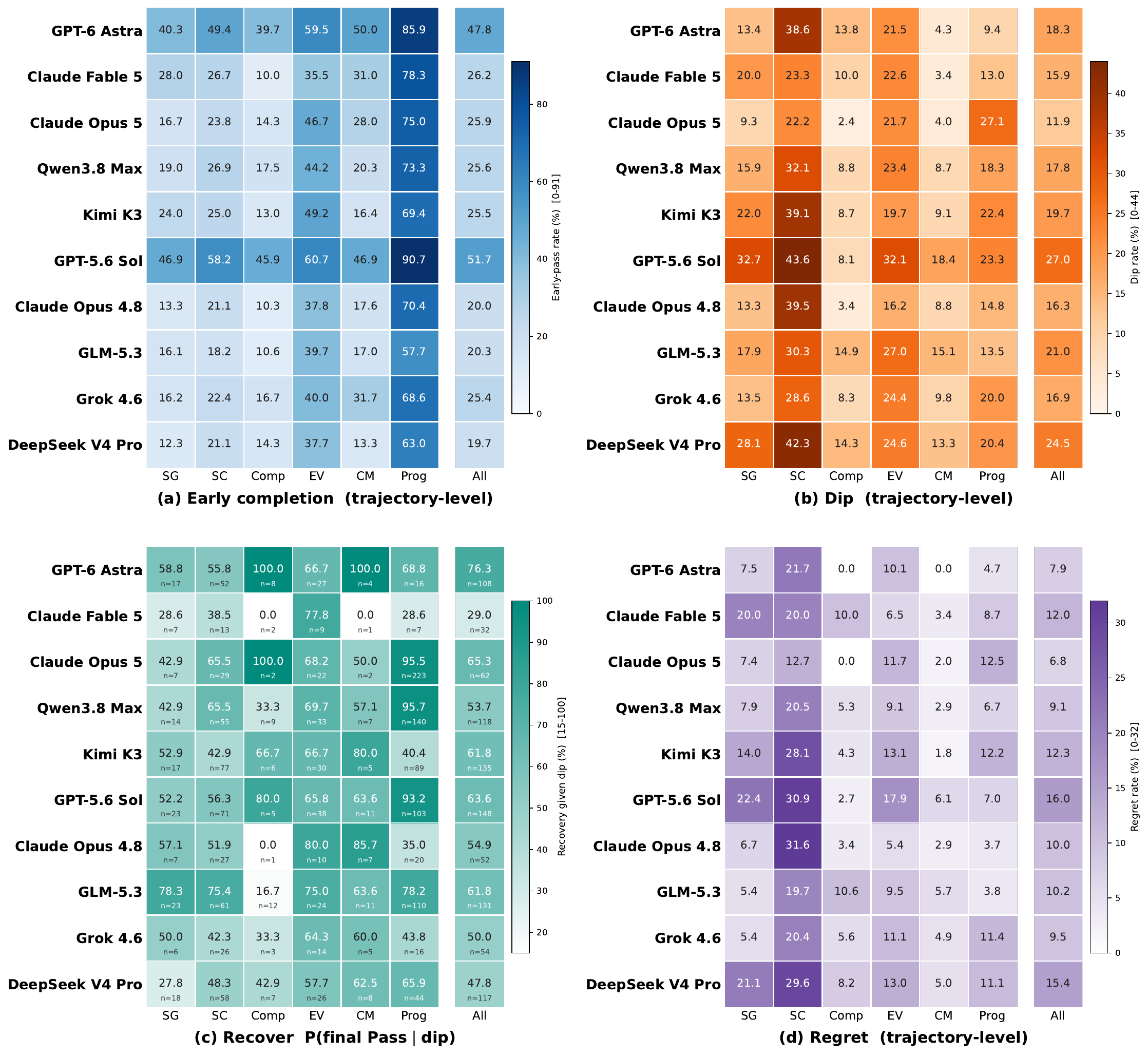}
    \caption{\textbf{Trajectory behavior across rubric dimensions and the programmatic
    channel, for all ten evaluated agents.} Rows are ordered by overall benchmark score.
    This full-population view extends the six-model main-text Figure~\ref{fig:behaviors_res} to the four agents
    with smaller judged samples (\claudefable{}, \claudeopuseight{}, \grok{}, and
    \deepseek{}).}
    \label{fig:regret_cat_heatmap_10model}
\end{figure*}

\begin{table*}[t]
\centering
\small
\setlength{\tabcolsep}{2.5pt}

\begin{tabular}{lccccc}
\toprule
\textbf{Task type} & \textbf{$n$} & \textbf{Early} &
\textbf{Dip} & \textbf{Recover} & \textbf{Regret} \\
\midrule
End-to-end analysis   & 134 & 36.2\% & 12.2\% & 10.3\% & 3.7\% \\
Literature synthesis  & 71 & 29.9\% & 6.6\% & 4.2\% & 3.3\% \\
Hypothesis validation & 117 & 20.9\% & 6.1\% & 4.5\% & 2.9\% \\
Tool orchestration    & 112 & 18.7\% & 8.3\% & 6.8\% & 3.3\% \\
Result interpretation & 47 & 13.3\% & 4.4\% & 2.1\% & 2.7\% \\
Code generation       & 112 & 8.6\% & 3.7\% & 2.1\% & 1.9\% \\
Error diagnosis       & 26 & 3.1\% & 1.5\% & 1.0\% & 0.8\% \\
\bottomrule
\end{tabular}

\caption{
Trajectory analysis by scientific task types.
}
\label{tab:regret_type}

\end{table*}

Beyond the aggregate trajectory statistics in the main text, we provide
additional analyses of how solution quality evolves within individual runs,
across models and evaluation dimensions, and across scientific task types.

\paragraph{Individual trajectories reveal strongly non-monotonic progress.}
Figure~\ref{fig:exemplar_combined}(a) shows representative trajectories from
different models and tasks. Their dynamics vary substantially: some runs reach
most of their eventual quality within only a few snapshots, while others remain
nearly unchanged for much of the trajectory before improving sharply near the
end. Several trajectories also exhibit temporary regressions, where previously
achieved quality is lost and later recovered. These examples illustrate why
endpoint scores provide an incomplete account of agent execution: similar final
quality can emerge through very different paths.

The same variation appears even when the task is held fixed.
Figure~\ref{fig:exemplar_combined}(b) compares six models on the same scientific
task. Some models establish executable progress early, whereas others improve
only after many additional interactions; the timing of improvements in the
programmatic and LLM-judge channels also differs substantially. Thus,
trajectory differences reflect not only task difficulty, but also how different
agents approach the same underlying problem.

\paragraph{Progress speed and stability vary independently across models.}
Figure~\ref{fig:regret_cat_heatmap_10model} extends the main-text analysis to
all ten evaluated agents. Early completion varies considerably across models:
\gptsol{} and \gptastra{} have the highest overall rates (51.7\% and 47.8\%),
whereas several other models remain near 20--26\%. High early completion,
however, does not necessarily imply stable subsequent execution. \gptsol{},
for example, combines the highest early-completion rate with substantially
higher dip (27.0\%) and regret (16.0\%) than \gptastra{} (18.3\% and 7.9\%).
The two models therefore reach useful states early but differ in how reliably
those gains are preserved.

The dimension-level results further show that regression is concentrated in
scientific content rather than presentation. Scientific Correctness has the
highest dip rates for most models and frequently the highest regret rates,
whereas Communication is consistently more stable. Programmatic criteria,
despite often being satisfied early, can also regress during later execution.
Together, these patterns show that agent progress is neither uniformly paced
nor monotonic: reaching a strong intermediate state and preserving it are
distinct aspects of execution behavior.

\paragraph{Trajectory dynamics also depend strongly on the scientific task type.}
Table~\ref{tab:regret_type} reveals a clear difference between open-ended
analysis and more localized technical tasks. End-to-End Analysis has the
highest early-completion rate (36.2\%) but also the highest dip rate (12.2\%),
indicating that substantial progress is often made early while later steps
continue to modify previously established results. Literature Synthesis also
shows relatively high early completion (29.9\%), but with fewer subsequent
regressions.

In contrast, Code Generation and Error Diagnosis rarely reach near-final
quality early (8.6\% and 3.1\%, respectively) and exhibit the lowest dip and
regret rates. This does not necessarily indicate easier or more stable
execution: rather, because successful criteria are established later, there
are fewer opportunities for previously achieved gains to be lost. Across task
types, early completion and regression therefore need to be interpreted
jointly rather than as independent indicators of agent quality.

Overall, the trajectory analysis exposes two properties hidden
by final scores: \emph{when} solution quality emerges and \emph{whether} it is
preserved. These properties vary systematically across models, evaluation
dimensions, and scientific task types, providing a complementary view of agent
execution beyond endpoint performance.

\section{Case Study: When execution and communication decouple}
\label{app:decoupling-case}

\sciswe{} scores every submission on two axes that measure different things. The
\emph{programmatic} axis re-executes the agent's artifacts and checks the numbers:
an independent reference checker recomputes the pipeline from the raw inputs,
compares the \texttt{answer.json} and every intermediate table element-wise at
tight tolerance, and reruns the pipeline on perturbed inputs so that hard-coded
answers fail. The \emph{LLM-judge} axis instead reads the scientific report
against a fixed rubric and asks whether the science was \emph{explained}
correctly---whether the method is framed properly, the diagnostics interpreted,
the hand-offs traced, and the conclusion justified---without re-checking any
number. The final score blends them,
$0.55\cdot\textrm{prog}+0.45\cdot\textrm{judge}$.

The two axes are complementary and, we argue, inseparable: a correct scientific
deliverable must both compute the right result \emph{and} explain why it is
right, and neither axis can stand in for the other. This appendix supports that
claim with two worked cases. Case~1 shows that the axes emphasize genuinely
different things, so collapsing them into one number is biased---a single model
can look perfect on one axis while failing the task on the other. Case~2 shows
that the axes are also what lets the benchmark separate model \emph{capabilities}:
different models fail on different axes, and a single score would silently merge
those failure modes.

\subsection{Case 1: one axis alone is biased}
\label{app:decoupling-single}

\begin{sloppypar}
Consider \textbf{cs-expansion-\allowbreak tool-\allowbreak orchestration-02} (domain \emph{CS Expansion},
task type \emph{tool orchestration}). The agent must select a budget-limited
labeled vertex set that maximizes the Graph Label Selection objective $\Psi(L)$,
following John et al.\ (2026), whose CC-BY PDF is bundled offline. No single
library call suffices: correctness flows through three load-bearing
hand-offs---(i) a hierarchical spectral decomposition producing a binary tree and
leaf$\rightarrow$vertex map, (ii) a $\tau$-search / tree-DP leaf-label selection
stage, and (iii) evaluation of $\Psi(L)$ on the \emph{original} graph. The agent
delivers \texttt{orchestrate.py} and \texttt{gls\_tools.py}, per-instance
intermediate hand-offs, and an \texttt{answer.json}. The programmatic axis has
eight weighted checks (total weight 129) that recompute every per-instance
labeled set and $\Psi$ value and rerun the pipeline on a perturbed input; the
judge rubric has 17 items (weight 61) spanning Scientific Correctness, Execution
\& Validation, Source Grounding, Completeness, and Communication.
\end{sloppypar}

\begin{sloppypar}
For \mbox{\gptsol{}}, the report passes \emph{all 17} rubric items (judge
$1.00$): it correctly describes the three-stage method, grounds it in the source
paper, and explains $\Psi$ and the hand-offs. Yet the programmatic axis is only
$0.10$---the model earns just 13 of 129 weight. Table~\ref{tab:decoupling-single}
shows why: only the small structural ``glue'' checks pass, while \emph{every}
result-bearing check fails. The per-instance labeled sets and $\Psi$ values do
not match the reference (\texttt{primary\_instances}, weight 52), the tree-DP
feasibility stage is wrong (\texttt{selection\_dp\_feasibility}, weight 30), the
hierarchical bisection is wrong (\texttt{spectral\_partitions}, weight 6), and
the pipeline fails to recompute correctly on a perturbed input
(\texttt{data\_dependent\_rerun}, weight 28).
\end{sloppypar}

\begin{table}[t]
\centering
\footnotesize
\setlength{\tabcolsep}{5pt}
\adjustbox{max width=\columnwidth}{%
\begin{tabular}{@{}lcc@{}}
\toprule
Programmatic check & Wt. & Result \\
\midrule
\texttt{primary\_instances}          & 52 & \textbf{fail} \\
\texttt{selection\_dp\_feasibility}  & 30 & \textbf{fail} \\
\texttt{data\_dependent\_rerun}      & 28 & \textbf{fail} \\
\texttt{spectral\_partitions}        &  6 & \textbf{fail} \\
\texttt{code\_rerun\_matches\_answer} &  4 & pass \\
\texttt{sparsifier\_structure}       &  3 & pass \\
\texttt{selection\_mapping\_handoff} &  3 & pass \\
\texttt{psi\_witness\_consistency}   &  3 & pass \\
\midrule
\multicolumn{2}{@{}l}{Earned / total weight} & 13 / 129 \\
\bottomrule
\end{tabular}}
\caption{\mbox{\gptsol{}} on \textbf{cs-expansion-tool-orchestration-02}:
judge $1.00$ (17/17 rubric items) but programmatic $0.10$. Only the small
structural ``glue'' checks pass; every result-bearing check fails---the
per-instance labeled sets and $\Psi$ values (\texttt{primary\_instances}), the
tree-DP selection (\texttt{selection\_dp\_feasibility}), the hierarchical
bisection (\texttt{spectral\_partitions}), and the perturbed-input rerun
(\texttt{data\_dependent\_rerun}).}
\label{tab:decoupling-single}
\end{table}

A judge-only metric would record this submission as flawless, when in fact its
central scientific result---the labeled sets and their $\Psi$ values---is wrong.
The same near-$1.0$-judge / near-$0.1$-programmatic split recurs for several
other models on this task, so it is a property of the submissions, not a single
noisy run. This is the sense in which a single score is biased: the two axes
reward different things, and whichever one is dropped, a whole class of failures
becomes invisible.

\subsection{Case 2: different models fail on different axes}
\label{app:decoupling-split}

\begin{sloppypar}
The second case shows the flip side: the two axes are what let the benchmark
separate models whose weaknesses lie in opposite places. The task is
\texttt{physics-expansion-tool-orchestration-03} (domain \emph{Physics
Expansion}, task type \emph{tool orchestration}): decide which finite-size
Floquet phase label best explains a twisted, periodically kicked six-qubit
spin-ring protocol, with a reproducible numerical pipeline rather than a
closed-form argument. The container is offline and ships Qiskit, SciPy, NumPy,
pandas, and NetworkX. Evidence is routed through four load-bearing stages: Qiskit
builds the two Pauli-operator Hamiltonians $H_z,H_x$; SciPy diagonalizes the
Floquet unitary $U=e^{-i\tau_x H_x}e^{-i\tau_z H_z}$, audits the ordered against
the reversed product, and solves a one-to-one eigenstate assignment; NetworkX
checks connectivity of the tracked-state graphs; and pandas scores the tracked
observables against candidate phase rules and reruns the sensitivity scenarios.
The programmatic axis recomputes all of this and compares the five intermediate
tables and \texttt{answer.json} element-wise at \texttt{rtol} $\approx$ \texttt{2e-6};
the judge rubric has 23 items (weight 90) dominated by Scientific Correctness and
Execution \& Validation.
\end{sloppypar}

Table~\ref{tab:decoupling-split} reports both axes for two capable models. Their
scores are near-perfect on one axis and near-zero on the other, in opposite
directions.

\begin{table}[t]
\centering
\footnotesize
\setlength{\tabcolsep}{4pt}
\adjustbox{max width=\columnwidth}{%
\begin{tabular}{@{}lccl@{}}
\toprule
Model & Prog. & Judge & Failure mode \\
\midrule
\deepseek{}   & \textbf{1.00} & 0.43 & right numbers, thin report \\
\claudeopusfive{} & 0.11 & \textbf{1.00} & flawless report, wrong numbers \\
\bottomrule
\end{tabular}}
\caption{The same task scored on both axes. A programmatic-only ranking places
\deepseek{} far ahead; an LLM-judge-only ranking places \claudeopusfive{} far ahead.
The two single-axis orderings are inverted.}
\label{tab:decoupling-split}
\end{table}

\paragraph{\deepseek{}: right numbers, thin report.} Its \texttt{answer.json},
all five intermediate tables, and the sensitivity rerun reproduce the reference
computation element-wise, so the programmatic axis is a perfect 1.00. The report,
however, passes only 10 of 23 rubric items. The failures are precisely the items
that demand physical interpretation: how eigenphases become the spectral
evidence, near-$\pi$ quasienergy membership as a finite-size diagnostic, entropy
/ imbalance / $X$-parity as distinct diagnostics, the role of the boundary twist,
the meaning of the selected winding index, why one-to-one tracking matters, and
\emph{both} Completeness hand-off items. The model computed everything correctly
but barely explained \emph{why}, so the judge scores it 0.43.

\paragraph{\claudeopusfive{}: flawless report, wrong numbers.} The report passes all
23 rubric items (judge 1.00), and the model even selects the correct phase label.
Yet the computation is wrong. As Table~\ref{tab:decoupling-answer} shows, three
of the five answer keys match the reference, but the selected-to-runner-up margin
is off by roughly $20\times$ and the winding index diverges in its fifth
significant digit (outside tolerance). Every recomputed intermediate table and
the data-dependent rerun fail, so the programmatic axis is only 0.11. The
conclusion is right and the narrative is airtight, but the numerical evidence
that is supposed to support them does not reproduce---a failure the report-only
judge cannot see.

\begin{table}[t]
\centering
\footnotesize
\adjustbox{max width=\columnwidth}{%
\begin{tabular}{@{}lcc@{}}
\toprule
\texttt{answer.json} key & \claudeopusfive{} & Reference \\
\midrule
\texttt{selected\_phase\_id}    & \multicolumn{2}{c}{\texttt{trivial\_prethermal\_bulk} (match)} \\
\texttt{selected\_drive\_scale} & \multicolumn{2}{c}{$0.82$ (match)} \\
\texttt{selected\_pi\_fraction} & \multicolumn{2}{c}{$0.046875$ (match)} \\
\texttt{selected\_score\_margin}  & $0.10033$ & $0.00483$ \\
\texttt{selected\_winding\_index} & $0.0015729$ & $0.0015928$ \\
\bottomrule
\end{tabular}}
\caption{\claudeopusfive{}'s answer against the reference. The headline phase label
is correct, but the supporting margin ($\sim$$20\times$ too large) and winding
index fall outside tolerance, and none of the five intermediate tables
reproduce.}
\label{tab:decoupling-answer}
\end{table}

Here \deepseek{} loses entirely on \emph{communication} and \claudeopusfive{} loses
entirely on \emph{execution}. A programmatic-only metric would rank \deepseek{}
well above \claudeopusfive{} (1.00 vs.\ 0.11); an LLM-judge-only metric would rank
them the other way (0.43 vs.\ 1.00). Reporting both axes separately, rather than
only as a blend, is what lets \sciswe{} distinguish ``computed it but can't
explain it'' from ``explained it but didn't compute it''---two failure modes that
a scalar score would silently merge.

\clearpage

\section{An Example of Task Package}\label{app:full-case}

To make the task package fully concrete, this appendix reproduces one case end to end: its
environment, its verbatim instruction, its reference solution, and both halves of the LLM rubric and the programmatic checker. The case is an End-to-End Analysis task \textit{bioinformatics-end-to-end-analysis-02}: select a
metagenomic assembly-and-binning pipeline that is robust across four contig-level stress axes,
under a precision-gated maximin policy grounded in the mounted MAG-E paper.

\begin{table*}[t]
\centering
\begin{tcolorbox}[
    enhanced,
    colback=white,
    colframe=purple!55,
    boxrule=1pt,
    arc=8pt,
    left=9pt,
    right=9pt,
    top=8pt,
    bottom=8pt,
    title={\textbf{Task package}},
    colbacktitle=purple!10,
    coltitle=black
]
\small
\renewcommand{\arraystretch}{1.3}
\setlength{\tabcolsep}{8pt}

\begin{tabularx}{\linewidth}{
    |>{\raggedright\arraybackslash
       \bfseries\color{purple!85!black}}p{0.30\linewidth}
    |>{\raggedright\arraybackslash}X|
}
\hline
Package files
& \textbf{Scientific and evaluation role} \\
\hline

\texttt{instruction.md}
& Verbatim research question, required analysis, execution command,
artifact contract, and answer schema. \\
\hline

\texttt{environment/}
& Frozen offline Docker environment with pinned dependencies. \\
\hline

\texttt{data/}
& One paper and fourteen visible data and quality-control files. \\
\hline

\texttt{solution/}
& Expert implementation, six output artifacts, report, and calibrated
reference decision. \\
\hline

\texttt{tests/rubric.json}
& Twenty observable scientific-reasoning criteria with a total weight
of 70. \\
\hline

\texttt{tests/}\newline
\texttt{check\_programmatic.py}
& Eight weighted checks, including a held-out replay in which every
identifier is relabeled. \\
\hline

\texttt{task.toml}
& Fixed 0.50 programmatic and 0.50 rubric-based score aggregation. \\
\hline
\end{tabularx}
\end{tcolorbox}

\caption{Contents of an example task package and the role of each component.}
\label{tab:task-package}
\end{table*}

\subsection{Instruction (\texttt{instruction.md}, verbatim)}\label{app:full-case-instr}
The instruction fixes the scientific goal, the exact rerun command, the authoritative source
sections, the numbered analysis, the deliverable contract, and the \texttt{answer.json} schema. It
names the two ``obvious'' leaders (whole-genome F-score and truth precision) only to require the
agent to explain why \emph{neither} is selected.

\begin{lstlisting}[style=taskmd]
# Contig-outlier-aware selection of a metagenomic assembly-and-binning pipeline

> **Execution mode:** Run this task end-to-end autonomously. Do not stop to ask for confirmation,
> do not print a draft for the user to run manually, and do not propose plans without executing
> them. Write every final artifact to `/home/agent/results/`. Your turn ends only when every
> deliverable is on disk.

Your implementation must rerun with:

```text
python code/mage_pipeline_analysis.py --data-dir /home/agent/data --output-dir /home/agent/results
```

## Background and research question

The mounted MAG-E paper evaluates complete assembly-and-binning pipelines against constructed
ground truth. Its contig-level analysis shows that ordinary whole-genome summaries can conceal
failures on contigs whose coverage or tetranucleotide composition differs from the originating
genome, as well as on prophage and shared sequence.

For this reduced deterministic benchmark, identify the precision-qualified pipeline with the most
robust recovery across four scientific axes:

- high coverage-distance contigs;
- high tetranucleotide-composition-distance contigs;
- prophage contigs; and
- shared contigs.

Return the selected pipeline and its limiting axis as one compositional decision. Explain why the
whole-genome F-score leader and the truth-precision leader are not selected.

Use only the mounted paper and supplied files. The paper's sections `The MAG-E simulator`, `The
MAG-E evaluator`, and `Contig-level analyses` are authoritative for the scientific definitions.
This fixture juxtaposes independent outputs from two assemblers, so it applies the paper's
contig-level calculations separately within each assembler before comparing complete pipelines.
That assembler partition, the precision prerequisite, and the maximin selection policy are
task-specific adaptations; they are not universal recommendations from the paper.

## Inputs

All inputs are in `/home/agent/data/`. Exact schemas, keys, units, and fixture quirks are documented
in `data_dictionary.md`.

| File | Scientific role |
| --- | --- |
| `pipelines for metagenome.pdf` | Sole primary source paper. |
| `sample_metadata.csv`, `species_profile.csv`, `strain_candidates.csv`, `strain_abundance_weights.csv`, `genome_catalog.csv` | Reconstruct the sample-specific truth manifest and expected coverage. |
| `pipeline_catalog.csv`, `contigs.csv`, `contig_genome_mappings.csv`, `bin_assignments.csv` | Reconstruct representative bins and base-pair recovery. |
| `contig_coverage_profiles.csv` | Six-sample coverage vector for every assembler contig. |
| `contig_tn_counts.csv` | Raw counts for the 256 tetranucleotides of every contig. |
| `genome_tn_profiles.csv` | Normalized 256-component tetranucleotide profile for each candidate genome. |
| `data_validation.json` | Machine-readable integrity and relationship audit for the exact mounted CSV bytes; provenance/QC only, not a substitute for analysis. |

Identifiers and row order are arbitrary and carry no ranking information. A contig may map to more
than one truth genome. `unbinned` is not a bin.

## Required analysis

1. Recover and implement the manifest conventions from `The MAG-E simulator`: the
   potential-strain screen, the choice within each strain cluster, the no-potential-strain
   fallback, and the allocation of the supplied realized weights. Compute expected coverage as
   sequenced bases x strain relative abundance / genome length. Keep every manifest row, but
   restrict pipeline evaluation to genomes with expected coverage at least `2.5x`.

2. Recover from `The MAG-E evaluator` how a representative bin is selected and how its
   whole-genome recall and truth precision are defined. Apply the same source-defined evaluation
   circuit to the prophage and shared-contig strata. Recover the treatment of multi-mapped contigs
   from the paper's `Contig-level analyses` section.

3. For every contig-to-genome mapping row, recover and implement the paper's coverage-distance and
   tetranucleotide-composition-distance definitions. Apply each definition within the mapping
   row's assembler, so the coverage reference population for this fixture is an
   `(assembler, genome_accession)` population across samples. The supplied coverage vectors, raw
   tetranucleotide counts, and genome profiles are observations for those calculations; do not
   replace them with scalar coverage, GC content, or a sample-wide profile.

4. Recover the paper's base-pair midpoint percentile convention and its scientific weighting.
   Apply it separately to the mapping-row population from each assembler. The acceptance-bearing
   exact-boundary convention is strict-lower/inclusive-upper, so an exact boundary stays in the
   preceding percentile. Break exact distance ties by `contig_id`, then `genome_accession`, for
   deterministic replay. Define each high-distance outlier stratum as percentile `>= 80`.

5. Within each pipeline/sample, aggregate mapped base pairs, not contig counts or macro means,
   for the four recovery axes. For an axis, recall is mapped base pairs assigned to that genome's
   representative bin divided by all mapped base pairs in that stratum. Prophage and shared axes
   use `region_class`; coverage and composition axes use their respective percentile. Separately
   macro-average truth precision and whole-genome recall over eligible genomes within each sample.

6. Apply this fixed policy without rounding intermediate values:

   - a pipeline is eligible only if sample-level macro truth precision is `>= 0.90` in every
     evaluable sample;
   - for each pipeline, average each of the four base-pair recovery axes across evaluable samples;
   - define `robust_recovery_minimum` as the smallest of those four means and name its
     `limiting_axis` as `coverage_outlier`, `composition_outlier`, `prophage`, or `shared`;
   - among eligible pipelines, maximize `robust_recovery_minimum`;
   - exact ties prefer higher global macro whole-genome recall, then higher minimum sample truth
     precision, then lexicographically smaller `pipeline_id`;
   - assign `policy_rank` as a one-based rank over precision-eligible pipelines in that policy
     order, and leave `policy_rank` empty for ineligible pipelines. The fixture has no unresolved
     tie.

7. Separately identify the global whole-genome F-score leader, where F-score is the harmonic mean
   of global macro truth precision and global macro whole-genome recall, and the global truth-
   precision leader. Explain each alternative's observable limiting-axis failure under the policy.
   Also identify every precision-ineligible pipeline and the sample or samples that fail the floor.

Use full-precision calculations. Numeric output is checked with absolute or relative tolerance
`1e-6`.

## Deliverables

Write all files below under `/home/agent/results/`.

| Path | Required contents |
| --- | --- |
| `code/mage_pipeline_analysis.py` | Readable executable implementation of the complete analysis. |
| `manifest.csv` | One row per reconstructed manifest genome. Required: `sample_id`, `species_cluster`, `strain_cluster`, `genome_accession`, `genome_length_bp`, `strain_relative_abundance`, `expected_coverage_x`, `analysis_eligible`. |
| `representative_bins.csv` | Key `pipeline_id`, `sample_id`, `genome_accession`. Required: `expected_coverage_x`, `representative_bin`, `truth_precision`, `whole_genome_recall`, `prophage_assembly_recall`, `shared_assembly_recall`. |
| `mapping_outlier_metrics.csv` | Key `assembler`, `contig_id`, `genome_accession`. Required: `sample_id`, `mapped_bp`, `region_class`, `contig_length_bp`, `coverage_distance`, `coverage_distance_percentile`, `tn_distance`, `tn_distance_percentile`. |
| `sample_pipeline_summary.csv` | Key `pipeline_id`, `sample_id`. Required: `n_units`, `mean_truth_precision`, `mean_whole_genome_recall`, `coverage_outlier_recall`, `composition_outlier_recall`, `prophage_recall`, `shared_recall`, `precision_floor_met`. |
| `pipeline_summary.csv` | Key `pipeline_id`. Required: `assembler`, `binner`, `binning_mode`, `mean_truth_precision`, `mean_whole_genome_recall`, `mean_genome_fscore`, `minimum_sample_truth_precision`, the four `mean_*_recall` axes, `robust_recovery_minimum`, `limiting_axis`, `precision_eligible`, and `policy_rank`. |
| `answer.json` | Structured headline decision using the schema below. |
| `report.md` | Scientific write-up with Context, Data/QC, Methods, Results, Validation, Limitations, and Conclusion. Explain the source-derived choices and why they matter, the two requested leaders, and each precision-ineligible pipeline with its failing sample(s). Identify the coverage and tetranucleotide inputs as reduced semi-synthetic fixture observations, distinguish the fixture policy from the paper's conclusions, and do not present the selected pipeline as a universal paper recommendation. |

Additional CSV columns are allowed. Booleans may be case-insensitive `true`/`false`; ineligible
`policy_rank` values must be empty.

`answer.json` must contain:

{
  "dataset_summary": {
    "n_samples": 0, "n_manifest_genomes": 0, "n_eligible_sample_genome_units": 0,
    "n_pipelines": 0, "coverage_threshold_x": 2.5, "truth_precision_floor": 0.9,
    "outlier_percentile_start": 80
  },
  "manifest_summary": {
    "n_species_without_potential_strain": 0, "n_genomes_below_coverage_threshold": 0
  },
  "selected_pipeline": {
    "pipeline_id": "string", "assembler": "string", "binner": "string", "binning_mode": "string",
    "mean_truth_precision": 0.0, "mean_whole_genome_recall": 0.0,
    "minimum_sample_truth_precision": 0.0, "mean_coverage_outlier_recall": 0.0,
    "mean_composition_outlier_recall": 0.0, "mean_prophage_recall": 0.0, "mean_shared_recall": 0.0,
    "robust_recovery_minimum": 0.0,
    "limiting_axis": "coverage_outlier|composition_outlier|prophage|shared",
    "margin_over_runner_up": 0.0
  },
  "alternative_summary": {
    "whole_genome_fscore_leader_id": "string", "whole_genome_fscore_leader_score": 0.0,
    "truth_precision_leader_id": "string", "truth_precision_leader_score": 0.0,
    "runner_up_pipeline_id": "string", "runner_up_limiting_axis": "string",
    "runner_up_robust_recovery_minimum": 0.0
  },
  "conclusion": {
    "decision": "<PIPELINE_ID>_<UPPERCASE_LIMITING_AXIS>", "selected_pipeline_id": "string",
    "limiting_axis": "string", "confidence": "high|medium|low",
    "primary_rationale": "string", "main_limitation": "string"
  }
}

In `conclusion.decision`, append the uppercase serialization of the selected lowercase
`limiting_axis` token to `selected_pipeline_id` with one underscore. Keep both standalone
`limiting_axis` fields in the lowercase enum shown above.

Before submitting, confirm that the CLI regenerates all artifacts, percentiles use the recovered
base-pair midpoint convention, the source-derived shared-contig handling is applied, every-sample
precision is applied before ranking, and all artifacts agree on the same pipeline and limiting axis.
\end{lstlisting}

\subsection{Environment}\label{app:full-case-env}
The task runs offline in a frozen Docker image. The visible workspace at \texttt{/home/agent/data/}
contains one source paper and fourteen data files (Table~\ref{tab:full-case-env}); no reference
artifact or checker input is mounted on the agent side.

\begin{table*}[h]
\centering\small
\setlength{\tabcolsep}{6pt}\renewcommand{\arraystretch}{1.25}
\begin{tabular}{@{}>{\raggedright\arraybackslash}p{3.0cm}>{\raggedright\arraybackslash}p{9.6cm}@{}}
\toprule
\textbf{Role} & \textbf{Files} \\
\midrule
Source \& docs &
\texttt{pipelines for metagenome.pdf}, \texttt{data\_dictionary.md},
\texttt{data\_validation.json} \\
Truth / manifest &
\texttt{sample\_metadata.csv}, \texttt{species\_profile.csv}, \texttt{strain\_candidates.csv},
\texttt{strain\_abundance\_weights.csv}, \texttt{genome\_catalog.csv} \\
Bins \& recovery &
\texttt{pipeline\_catalog.csv}, \texttt{contigs.csv}, \texttt{contig\_genome\_mappings.csv},
\texttt{bin\_assignments.csv} \\
Contig matrices &
\texttt{contig\_coverage\_profiles.csv}, \texttt{contig\_tn\_counts.csv},
\texttt{genome\_tn\_profiles.csv} \\
\bottomrule
\end{tabular}
\caption{The files mounted at \texttt{/home/agent/data/}.}
\label{tab:full-case-env}
\end{table*}

\subsection{Reference solution}\label{app:full-case-solution}
The package ships an expert reference solution (\texttt{solution/}): the analysis code
(\texttt{code/mage\_pipeline\_analysis.py}), all six required CSV/JSON artifacts, and the write-up
(\texttt{report.md}, with Context / Data \& QC / Methods / Results / Validation / Limitations /
Conclusion). Its headline \texttt{answer.json} --- the reference decision the checker is calibrated
against --- is reproduced below. The selected pipeline is \texttt{MAGE-83}, limited by the
\emph{composition-outlier} axis; note that this is \emph{neither} the whole-genome F-score leader
(\texttt{MAGE-47}) \emph{nor} the truth-precision leader (\texttt{MAGE-31}) --- the two decoys the
instruction asks the agent to reject.

\begin{lstlisting}[style=taskfile]
{
  "selected_pipeline": {
    "pipeline_id": "MAGE-83", "assembler": "metaSPAdes", "binner": "COMEBin",
    "binning_mode": "multi", "limiting_axis": "composition_outlier",
    "mean_truth_precision": 0.9875, "mean_whole_genome_recall": 0.3300,
    "minimum_sample_truth_precision": 0.9607,
    "mean_coverage_outlier_recall": 0.5147, "mean_composition_outlier_recall": 0.2862,
    "mean_prophage_recall": 0.6123, "mean_shared_recall": 0.4721,
    "robust_recovery_minimum": 0.2862, "margin_over_runner_up": 0.2375
  },
  "alternative_summary": {
    "whole_genome_fscore_leader_id": "MAGE-47", "whole_genome_fscore_leader_score": 0.5828,
    "truth_precision_leader_id": "MAGE-31",     "truth_precision_leader_score": 0.9988,
    "runner_up_pipeline_id": "MAGE-47", "runner_up_limiting_axis": "shared",
    "runner_up_robust_recovery_minimum": 0.0488
  },
  "conclusion": {
    "decision": "MAGE-83_COMPOSITION_OUTLIER", "selected_pipeline_id": "MAGE-83",
    "limiting_axis": "composition_outlier", "confidence": "high",
    "primary_rationale": "Clears the truth-precision floor in every evaluable sample and
       maximizes the minimum mean recovery across the four axes.",
    "main_limitation": "A deterministic reduced synthetic-community fixture, not a full
       replication of the paper's 4,800 tasks."
  },
  "dataset_summary": { "n_samples": 6, "n_manifest_genomes": 18,
    "n_eligible_sample_genome_units": 17, "n_pipelines": 8, "coverage_threshold_x": 2.5,
    "truth_precision_floor": 0.9, "outlier_percentile_start": 80 },
  "manifest_summary": { "n_species_without_potential_strain": 2,
    "n_genomes_below_coverage_threshold": 1 }
}
\end{lstlisting}

\subsection{Grading, part 1 --- the LLM rubric (\texttt{tests/rubric.json})}\label{app:full-case-rubric}
The judge grades \emph{scientific reasoning and source grounding}, not numeric values: its context
(\texttt{judge\_config.json}) states that ``exact identifiers and numeric values are owned by
\texttt{check\_programmatic.py}''. Each criterion is an observable pass/fail statement with named
\texttt{grader\_inputs} (the report, the code, and the specific CSV it must be consistent with); the
judge sees the source excerpts and the rubric but never the reference score. The complete rubric is
20 criteria over four categories, total weight 70 (Table~\ref{tab:full-case-rubric}).

\begin{table*}[h]
\centering
\setlength{\tabcolsep}{4pt}\renewcommand{\arraystretch}{1.15}
\begin{tabular}{@{}l c c >{\raggedright\arraybackslash}p{10.4cm}@{}}
\toprule
\textbf{\#} & \textbf{Category} & \textbf{Weight} & \textbf{Criterion} \\
\midrule
1  & SG & 4 & Explains why the paper's potential-strain screen is necessary before constructing the truth manifest. \\
2  & SG & 3 & Distinguishes the paper's within-strain-cluster choice from its no-potential-strain fallback. \\
3  & SG & 3 & Explains the source-defined per-genome representative-bin estimand. \\
4  & SG & 4 & Explains the paper's treatment of multi-mapped shared contigs as mapping-level evidence. \\
5  & SG & 4 & Explains why the paper uses base-pair midpoint percentiles for contig-level analysis. \\
6  & SC & 4 & Interprets how the expected-coverage eligibility threshold changes the evaluation population. \\
7  & SC & 4 & Interprets coverage distance as task-local assembler-and-genome-conditioned across-sample deviation. \\
8  & SC & 4 & Interprets tetranucleotide cosine distance as a composition-shift measure. \\
9  & SC & 4 & Explains why the four recovery axes are distinct scientific stress tests. \\
10 & SC & 4 & Explains the consequence of mapped-base aggregation for recovery. \\
11 & SC & 4 & Justifies the every-sample precision prerequisite as a non-compensatory safeguard. \\
12 & SC & 4 & Interprets the maximin policy as protection against a weakest-axis failure. \\
13 & SC & 4 & Explains the selected pipeline's limiting axis as the decision bottleneck. \\
14 & SC & 4 & Explains why the whole-genome F-score alternative loses under the declared question. \\
15 & SC & 4 & Explains why the truth-precision leader is not the robust-recovery choice. \\
16 & EV & 3 & Uses sample-level evidence to explain every precision-ineligible pipeline. \\
17 & EV & 3 & Validates the distinction between every-sample eligibility and global precision leadership. \\
18 & EV & 2 & Justifies the task-local assembler partition without presenting it as a source convention. \\
19 & CM & 2 & Scopes the pipeline conclusion to the reduced semi-synthetic task fixture. \\
20 & CM & 2 & Must \emph{not} claim perturbation, ablation, or sensitivity results absent from the submitted artifacts. \\
\bottomrule
\end{tabular}
\caption{The complete 20-criterion rubric for the full case (total weight 70). Category codes:
SG = Source Grounding, SC = Scientific Correctness, EV = Execution \& Validation, CM = Communication.}
\label{tab:full-case-rubric}
\end{table*}

One criterion is shown verbatim to convey the JSON shape, including the calibration bar in its
\texttt{description} and the artifacts the judge is allowed to read:

\begin{lstlisting}[style=taskfile]
{ "id": 1, "category": "Source Grounding", "weight": 4,
  "criterion": "Explains why the paper's potential-strain screen is necessary before
     constructing the truth manifest.",
  "description": "Award only when the report explains that LOW effective lambda makes
     adjusted ANI unreliable; potential strains therefore require non-LOW effective lambda
     and adjusted ANI at least 99.8%, plus a 99.5% lower confidence bound when coverage is
     not HIGH. The explanation must connect those conditions to excluding unreliable strain
     matches before truth construction.",
  "source": "pipelines for metagenome.pdf, Methods, The MAG-E simulator",
  "grader_inputs": ["report.md", "code/mage_pipeline_analysis.py"] }
\end{lstlisting}

\subsection{Grading, part 2 --- the programmatic checker (\texttt{tests/check\_programmatic.py})}\label{app:full-case-prog}
The checker recomputes every deliverable from a golden bundle and, crucially, re-runs the agent's
own \texttt{mage\_pipeline\_analysis.py} on a \emph{held-out, relabeled} copy of the data: all
pipeline, genome, and bin identifiers are rewritten (\texttt{CAL-P01}, \texttt{CAL-G001},
\texttt{CAL-B0001}, \ldots) and the analysis must reproduce the correct decision under the new
labels. This \texttt{heldout\_pipeline\_replay} check carries the largest single weight (10 of 30),
so hard-coding \texttt{MAGE-83} or memorizing any identifier or metric table fails. We list the
eight named checks with their weights and the checker's function names --- not their bodies.

\begin{lstlisting}[style=taskpy]
# --- weighted check registry (CHECK_WEIGHTS; total weight 30) ------------------
CHECK_WEIGHTS = {
    "primary_selection":          4.0,   # selected pipeline + limiting axis match
    "manifest_science":           2.0,   # manifest reconstruction values
    "representative_bin_science": 3.0,   # per-genome representative-bin metrics
    "mapping_outlier_science":    4.0,   # coverage / tetranucleotide distances + percentiles
    "sample_recovery_science":    3.0,   # four base-pair recovery axes, per sample
    "pipeline_policy_science":    3.0,   # eligibility, maximin ranking, policy_rank
    "dataset_counts":             1.0,   # answer.json dataset/manifest summaries
    "heldout_pipeline_replay":   10.0,   # RERUN under relabeled IDs -> anti-hard-coding
}

# --- checker functions (names only) -------------------------------------------
def _load_csv(...): ...                  # I/O + golden-bundle loaders
def _load_bundle(...): ...
def _expected_mapping(...): ...
def _safe_check(...): ...                # tolerant numeric compare (RTOL=ATOL=1e-6)
def _close(...): ...
def _as_bool(...): ...
def _compare_table(...): ...

def _primary_selection(...): ...         # the eight weighted science checks
def _manifest_science(...): ...
def _representative_bin_science(...): ...
def _mapping_outlier_science(...): ...
def _sample_recovery_science(...): ...
def _pipeline_policy_science(...): ...
def _dataset_counts(...): ...
def _all_science(...): ...

def _heldout_case(data_dir, target): ... # build the relabeled (CAL-*) held-out copy
def _rewrite_csv(...): ...
def _heldout_pipeline_replay(results): ... # import + rerun agent code on held-out data

def check(results_dir): ...              # aggregate -> programmatic score in [0, 1]
def main(): ...
\end{lstlisting}

The final task reward is $0.50\cdot\text{(programmatic score)} + 0.50\cdot\text{(rubric score)}$,
as declared in \texttt{task.toml} for this end-to-end-analysis task.

\end{document}